\documentclass{article}
\pdfoutput=1

\usepackage[utf8]{inputenc} % allow utf-8 input
\usepackage[T1]{fontenc}    % use 8-bit T1 fonts
\usepackage{hyperref}       % hyperlinks
\usepackage{url}            % simple URL typesetting
\usepackage{booktabs}       % professional-quality tables
\usepackage{amsfonts}       % blackboard math symbols
\usepackage{amssymb}        % blackboard math symbols
\usepackage{amsmath}        % blackboard math symbols
\usepackage{nicefrac}       % compact symbols for 1/2, etc.
\usepackage{microtype}      % microtypography
\usepackage{color}
\usepackage{graphicx}
\usepackage{subcaption}
\usepackage{wrapfig}
\usepackage{multirow}
\usepackage{soul}
\usepackage[top=3cm, bottom=3cm, left=3.5cm, right=3.5cm]{geometry}

\usepackage{pgfplots}
\usepackage{tikzscale}
\usepgfplotslibrary{fillbetween,groupplots}
\usetikzlibrary{matrix,positioning}

\usepackage[backend=biber,style=numeric]{biblatex}
\pgfplotsset{compat=newest}

\title{Anytime Neural Prediction via Slicing Networks Vertically}

\author{
Hankook Lee \thanks{School of Electrical Engineering, Korea Advanced Institute of Science and Technology, Daejeon, Korea (hankook.lee@kaist.ac.kr, jinwoos@kaist.ac.kr)} \and Jinwoo Shin \footnotemark[1]
}

\begin{document}

\maketitle

\begin{abstract}
The pioneer deep neural networks (DNNs) have emerged to be deeper or wider for improving their accuracy in various applications of artificial intelligence. However, DNNs are often too heavy to deploy in practice, and it is often required to control their architectures dynamically given computing resource budget, i.e., anytime prediction. While most existing approaches have focused on training multiple shallow sub-networks jointly, we study training thin sub-networks instead.
To this end, we first build many inclusive thin sub-networks (of the same depth) under a minor modification of existing multi-branch DNNs,
and found that they can significantly outperform the state-of-art dense architecture for anytime prediction.
This is remarkable due to their simplicity and effectiveness, but training many thin sub-networks jointly faces a new challenge on training complexity. To address the issue, we also propose a novel DNN architecture by forcing a certain sparsity pattern on multi-branch network parameters, making them train efficiently for the purpose of anytime prediction.
In our experiments on the ImageNet dataset, its sub-networks 
have up to $43.3\%$ smaller sizes (FLOPs) compared to those of the state-of-art anytime model with respect to the same accuracy.
Finally, we also propose an alternative task under the proposed architecture using a hierarchical taxonomy, which
brings a new angle for anytime prediction.
%for preventing the performance loss of thin sub-networks 
\end{abstract}

% !TEX root = main.tex

\section{Introduction}

Deep neural networks (DNNs) have demonstrated state-of-the-art performance on many
artificial intelligence
applications such as speech recognition \cite{deepspeech_hannun2014}, image
classification \cite{fastrcnn_girshick2015}, video prediction \cite{learning_villegas2017}
and medical diagnosis \cite{healthcare_caruana2015}. 
One of key components underlying their success is on advanced DNN architectures 
such as AlexNet \cite{alexnet_krizhevsky2012}, VGGNet \cite{vgg_simonyan2014}, 
Inception \cite{inception_szegedy2015}, %{FractalNet \cite{FractalNet_Larsson2016},}
ResNet \cite{resnet_he2016} and DenseNet \cite{densenet_huang2017}.
%These success are based on the
%improvement of neural network architectures
%such as AlexNet \cite{alexnet_krizhevsky2012}, VGGNet \cite{vgg_simonyan2014},
%Inception \cite{inception_szegedy2015}, ResNet \cite{resnet_he2016}, and DenseNet \cite{densenet_huang2017}.
%In particular, ResNet is the first network training very deep layers, 
%and DenseNet uses thousands of channels efficiently.
Although they were originally developed for image classification, %their underlying ideas
%are generic and 
they also have provided influential impacts on designing deep and wide network architectures
for other related tasks, e.g., 
{3D-object reconstruction \cite{choy20163d}, super-resolution \cite{ledig2016photo}, image compression \cite{theis2017lossy} and image generation \cite{sinha2017surfnet}}.

%By utilizing these architectures, one can achieve high performance on many tasks with very
%deep and wide models.

However, the heavy models are often not suitable for real-world applications due to
their resource budgets, e.g.,
limited memory in mobile devices, low latency for
%have limited memory, so
%they cannot use such models. Also, time-critical services such as
autonomous driving and real-time constraints for Internet video delivery.
%require low latency, thus they should use low-computation models. 
Motivated by this, extensive research efforts recently have been made
for designing light neural networks of high performance under various approaches, e.g.,
network pruning \cite{compression_han2015, pruning_li2016} and transfer
learning \cite{fitnet_romero2014, distillation_hinton2015}.
%many approaches have been studied such as pruning \cite{compression_han2015}
%and knowledge distillation \cite{distillation_hinton2015}.
However, they are primarily targeting a single 
%These approaches are effective under only one fixed 
resource budget constraint. Hence, if an
application should work under various resource budgets adaptively or the computing power of clients' device is heterogeneous,
then 
%a new proper model %for the application or the device 
%is required since
%the previous one cannot be used under lower budget or cannot achieve good
%performance under higher budget. In other words, new model should be trained
%for each new device, or 
all models of various sizes should be ready separately.
%for varying
%resource budget tasks. 
%It is too time-consuming to train and inefficient to store.
It is too inefficient to train and store.

For such applications, designing a single model which can operate under
various computational resource budgets dynamically is very important. The problem is often called
\textit{anytime/adaptive prediction} \cite{anytime_zilberstein1996, speedboost_grubb2012}.
%whose
%is a related
%task to predict labels of images under anytime. The ultimate goal of this task is to
%find a model which has state-of-the-art performance under any computational cost.
%goal is to design scalable/adaptive network architectures, i.e., a part of DNN 
%should bring an incremental benefit to use it.
%have good performance.
%as well as the solely optimized corresponding network.
% In the point of view, to design an architecture which be able to operate under
% different resource budgets is very important.
% Anytime prediction is a related
% task to predict labels of images under anytime.
% The ultimate goal of the task is to find a model which has state-of-the-art performance
% under any computational cost.
% The way to predict labels under varying budgets with neural networks is to use
% parts of the networks, i.e., \textit{sub-networks}.
% A sub-optimal goal for anytime prediction is that each sub-network has good performance 
% as well as the solely optimized sub-network.
Most recent works on this line
\cite{branchynet_teerapittayanon2016,dyrouting_mcgill17, adaptive_bolukbasi17, sact_figurnov2017,nestednet_kim2017,msdnet_huang2017}
use intermediate features of DNN, possibly attached by 
%The mainstream of anytime prediction is using intermediate features with 
additional
classifiers to produce multiple outputs at different layers. 
%This is for choosing a proper sub-network for faster inference on demand:
That is, shallower and deeper sub-networks are used under smaller and larger resources (but, lower and higher accuracy), respectively, on demand.
%Then, Training the original network and the
%\textit{shallow} sub-networks jointly makes anytime prediction be possible. 
%Since optimizing the corresponding multiple losses jointly is possible by only single forward and backward propagation passes, the anytime
%DNN models can be trained efficiently. 
%There are many works
%which utilize shallow sub-networks for anytime prediction. 
%After the joint-training, one can  %faster inference,
%while the full-network provides the maximum accuracy among all sub-networks.
However, joint-training all such sub-networks together is not easy since
%\sout{each layer should capture both coarse-level and high-level features.}
shallower ones cannot capture level features, and deeper ones
might be badly trained if one forces intermediate layers to produce the final outputs.
To tackle the challenge, % the issue, % use
multi-scale inputs/features \cite{dyrouting_mcgill17, msdnet_huang2017}
and dense connections \cite{msdnet_huang2017} have been used, but
shallow sub-networks should fundamentally suffer from lack of expressive capacity in feature representation.

\textbf{Contribution.}
\iffalse
Our main contribution is on providing experimental supports to claim that many recent
DNN architectures have potentials by themselves to achieve good anytime performance
using their own sub-networks. Specifically, we observe that they have multiple branches
at each block and consider their inclusive \textit{thin} sub-networks which obtained by
removing some branches of the full-network in a progressive manner, while keeping the
original depth. Since all sub-networks have the same depth, we can train all thin
sub-networks jointly without the similar issues caused by joint-training with shallow
ones.
\fi
To overcome the fundamental limitation on training shallow sub-networks jointly, 
we aim for doing \textit{thin} ones instead. 
%To achieve the high performance anytime prediction performance,
To this end, we first observe that many recent state-of-the-art DNN architectures of residual-type 
have multiple branches
at each block. We consider their inclusive thin sub-networks obtained by
removing some branches of the full-network in a progressive manner while keeping the
original depth.
Then, we train all sub-networks (including
the full-network) jointly under a minor modification:
each sub-network has independent batch normalization layers
\cite{bn_ioffe2015}. %\sout{ and shallow classifiers}.
This simple approach is promising as it is based on
existing multi-branch DNNs that have achieved
the state-of-art performance for the standard, non-anytime prediction task.
\iffalse
In particular, we perform experiments on 
%a multi-branch architecture,
ResNeXt
\cite{resnext_xie2017}
that is a natural
modification of the standard ResNet architecture %to have many more branches
and has achieved the state-of-the-art accuracy for both ImageNet
%\footnote{ResNeXt won the 2nd place in the ILSVRC 2016 competition.} 
\cite{ILSVRC15} 
and CIFAR-10/100 datasets \cite{cifar, Gastaldi17_shake}.
%has achieved the state-of-the-art performance
%currently used in many state-of-the-art architectures 
We emphasize that ResNeXt is originally proposed for improving the performance of ResNet
on 
the single task of image classification, 
while we use it for measuring the performance of all tasks of sub-networks under
their joint training.
\fi

Our first major founding is that the performance of the full-network, named I({nclusive})-ResNeXt under the joint training
is not degraded much compared to the original solely trained network without considering sub-networks.
In other words, I-ResNeXt is able to make each sub-network perform their own functionality of high performance.
%} of the full-network.
In our experiments, I-ResNeXt, significantly outperforms the state-of-art anytime prediction model, MSDNet \cite{msdnet_huang2017}
on CIFAR datasets. In particular, its sub-networks %and full-network have 
have up to $56.0\%$ and $48.7\%$ smaller sizes (FLOPs)
compared to those of MSDNet of the same accuracy on CIFAR-10 and CIFAR-100, respectively. 

However, the above simple approach has the following
drawback: training its $K$ thin sub-networks jointly requires $K$ independent (forward and backward) propagation passes
at each iteration, making the overall training procedure $K$ times slower. 
Here, we remark that 
joint-training all shallow networks, e.g., in MSDNet, requires
only a single propagation pass at each iteration due to their interruptible property.
%as inferring all of them requires only a single propagation pass.
%One problem of using thin sub-networks is that $O(K)$ propagation steps are required for
%joint-training with $K$ thin sub-networks since each network has distinct intermediate
%features. To overcome this disadvantage, 
To address the training issue, we propose a novel multi-branch architecture whose
thin sub-networks can be trained jointly using a single propagation pass at each iteration. 
{In particular, we 
design a sparse version of I-ResNeXt, named IS-ResNeXt by
enforcing a certain sparsity pattern of network parameters:}
%This results in that the $k$-th channel at some layer is computable only using the $1,2,\ldots,k$-th channels at lower layers,
%and propagation computations of all thin sub-networks can be amortized, similarly as the case of 
%those of shallow networks. 
%Forcing such sparse parameters might leads a slight accuracy, 
%but 
IS-ResNeXt is $K$ times faster to train, enabling us to apply it to large-scale datasets.
In our experiments, sub-networks of IS-ResNeXt %and full-network have 
have up to $43.3\%$ smaller sizes (FLOPs)
compared to those of MSDNet of the same accuracy on the ImageNet \cite{imagenet_deng2009} dataset.

{
Finally, we also try an alternative task, called hierarchical anytime prediction, 
assuming the hierarchical taxonomy is available.
%apply IS-ResNeXt to a new anytime prediction task, called by hierarchical anytime prediction,
In this case, 
a model is allowed to predict coarse-level labels in the taxonomy
%when only small budgets are allowed, otherwise does fine-level labels
for maintaining the original accuracy level across all resource budgets. 
%This is from the natural motivation
Namely, it is better to produce less-informative predictions rather than incorrect ones.
We show that the proposed IS-ResNext also works well for the new problem, 
e.g., the taxonomy information allows us to improve the worst-case accuracy $31.2\%\rightarrow 47.5\%$ of sub-networks
on the Caltech-UCSD Birds \cite{cub_wah2011} dataset.
While the improvement might be not surprising as we allow to lose the predictive information for small sub-networks,
this new task provides a new angle toward a practical anytime predictor and
show the robustness of our architectures under significantly more tasks to perform.

\section{Preliminaries}

\subsection{Anytime prediction under shallow neural sub-networks}\label{sec:pre:shallow}

We first describe the model of anytime prediction \cite{speedboost_grubb2012} and its training loss. % and define loss formally.
Let $f:\mathbf{x}\rightarrow\hat{y}$ be a model and $\tau(f)$ be its computational cost. The
cost function $\tau$ can be actual CPU/GPU time or the number of multiply-addition operations
(FLOPs). In this paper, we use FLOPs as the measure as it only depends on models (not devices). 
For given time $T$, let $f_{\langle T\rangle}:\mathbf{x}\rightarrow\hat{y}$
be a restriction of the model $f$ with $\tau(f_{\langle T\rangle})\le T$. One can say that
$f$ is an anytime predictor if $f_{\langle T\rangle}$ can produce an output of high quality for
any time
budget $T$. %, $f_{\langle T\rangle}$ achieves good performance.
There are several desirable properties for anytime predictors \cite{anytime_zilberstein1996}:
\textit{monotonicity} - the quality of predictions is non-decreasing over time budgets;
% \textit{diminishing returns} - the improvement of quality diminishes over time;
\textit{optimality} - the quality at any time budget is close to the optimal quality under it;
\textit{interruptibility} - the predictor can be stopped at any time with predictions.
Here, the interruptibility is useful when the time budget $T$ is unknown when an input $\mathbf{x}$ is ready to process.
However, in many practical scenarios, one can decide $T$ in advance, e.g.,
environments typically change slowly compared to $\tau(f)$ or the client's device information is available.
%the property is optional if $T$ is given before inference or the computing time
%$\tau(f)$ is short. Since neural networks can produce an output for single input within few
%milliseconds, we do not consider the property.

Let $f_1,f_2,\ldots,f_K$ be the possible restrictions of an anytime predictor $f$. Then,
for a given training dataset $\mathcal{D}$,
%=\{(\mathbf{x}_i,y_i):i=1,\ldots,N\}$,
the loss
for anytime prediction can be defined as the following:
\begin{equation}\label{eq:anytime-loss}
%\mathcal{L}_\text{anytime}(\mathcal{D},f)=\frac{1}{N}\sum_{i=1}^N\sum_{k=1}^K\mathcal{L}(y_i,f_k(\mathbf{x}_i))
\mathcal{L}_\text{anytime}(\mathcal{D},f)=\frac{1}{|\mathcal{D}|}\sum_{(\mathbf{x},y)\in\mathcal{D}}\sum_{k=1}^K\mathcal{L}(y,f_k(\mathbf{x})),
\end{equation}
where $\mathcal{L}$ is a loss (e.g., cross entropy) between ground-truth and model-prediction.

%\subsection{Shallow sub-networks for anytime prediction}\label{sec:pre:shallow}

In the case of a DNN $f$, it is natural to consider shallow sub-networks as the restriction
models $\{f_k\}$. Let $h^{(l)}$ be the $l$-th layer function and $\mathbf{x}^{(l)}$ be the input
of $h^{(l)}$, i.e., $\mathbf{x}^{(l+1)}=h^{(l)}(\mathbf{x}^{(l)})$. Then, one can produce
an output by using an intermediate feature $x^{(l_k+1)}$, i.e.,
$f_k(\mathbf{x})=g_k(\mathbf{x}^{(l_k+1)})$ where $g_k$ be an auxiliary output function (or classifier). In
this case, only the first $l_k$ layers are required to compute, i.e., the output can be produced from
a shallow sub-network.
Since the sub-network is more efficient to compute, one might hope to choose it
instead of the full-network for faster inference (but, potentially sacrificing accuracy) on
demand.
%Any DNN architectures can be used for anytime prediction with their shallow sub-networks.
However, obtaining good shallow sub-networks and the full-network together, i.e., training
them jointly sharing parameters, is fundamentally difficult because %of the following reasons:
(a) an auxiliary output function connected
to an intermediate layer under a shallow sub-network cannot capture both coarse-level and
high-level features and (b)
forcing the intermediate features to produce outputs might degrade performance of the full-network.
This phenomenon hurts the desired optimality of anytime prediction: the overall performance of shallow sub-networks are degraded
compared to the case
of training them separately/independently without sharing their parameters. We provide
experimental supports on this %performance drop and compare them with our different approach 
in Section
\ref{sec:res:ablation}.

% Let $f_{\langle T\rangle}$ is a restrict version for satisfying the time constraint $\tau(f|_T)\le T$.
% Then, the goal of anytime prediction task is to find a model $f$ such that $f_{\langle T\rangle}$ can produce accurate
% prediction $f|_T(\mathbf{x})$ for an input $\mathbf{x}$ under given anytime budget $T$.

% We assume that the model $f$ has only finite number $K$ of restricted versions $f_1,f_2,\ldots,f_K$.

% Given a training dataset $\mathcal{D}=\{(\mathbf{x}_i,y_i)~|~i=1,\ldots,N\}$ and
% a time budget stochastic distribution $\mathcal{T}$,
% we can formulate anytime prediction loss as following:
% \begin{equation*}
% \mathcal{L}_\text{anytime}(\mathcal{D},f)=\frac{1}{N}\sum_{i=1}^N\mathbb{E}_{T\sim\mathcal{T}}\left[\mathcal{L}(y_i,f_{\langle T\rangle}(\mathbf{x}_i))\right]
% \end{equation*}
% where $\mathcal{L}$ is a loss function between ground-truth and prediction such as cross entropy loss.

% If the model $f$ has only finite number $K$ of restricted versions $f_1,f_2,\ldots,f_K$, then one can
% simplify $\mathcal{L}_\text{anytime}$ as following:
% We use \eqref{eq:anytime-loss} for training models since neural networks have finite number of
% sub-networks.

\subsection{Multi-branch neural architectures}

For developing better anytime prediction models,
we focus on multi-branch DNN models which include many existing state-of-the-art architectures, e.g.,
Inception \cite{inception_szegedy2015}, FractalNet \cite{fractalnet_larsson2016}, 
ResNet \cite{resnet_he2016} and DenseNet \cite{densenet_huang2017}.
They commonly apply the following module/block, repeatedly for building deep models.
Let $\mathcal{F}_i$ {be} the $i$-th branch that can be any function, e.g., neural networks or
polynomials, and $\mathcal{A}$ is a function aggregating them, e.g., addition or
concatenation. For an input $\mathbf{x}$, the branches of a block process it independently,
and then the block aggregates them {to produce an output $\mathbf{y}$} as
\begin{align*}
\mathbf{y}=\mathcal{A}\left(\mathcal{F}_0(\mathbf{x}),\mathcal{F}_1(\mathbf{x}),\cdots,\mathcal{F}_C(\mathbf{x})\right).
\end{align*}
By using the output $\mathbf{y}$ as an input  of the next block iteratively,
deeper networks can be constructed.

One can immediately observe that ResNet is a two-branch architecture having the identity
branch $\mathcal{F}_0(\mathbf{x})=\mathbf{x}$ and the additive aggregation, i.e.,
$\mathbf{y}=\mathbf{x}+\mathcal{F}(\mathbf{x})$. The ResNeXt \cite{resnext_xie2017} architecture is an extension
of ResNet: the former has more explicit branches and often outperforms the
latter, e.g., ResNeXt achieved the 2nd place on the {Large-Scale Visual Recognition Challenges (ILSVRC)} 2016 classification task and state-of-the-art accuracy on CIFAR datasets
with additional regularization technique \cite{shake_gastaldi17}.
%state-of-the-art accuracy for both ImageNet \cite{imagenet_deng2009}
% and CIFAR-10/100 \cite{cifar_krizhevsky2009a, shake_gastaldi17} datasets.
ResNeXt uses a single identity branch as like ResNet and additional branches
$\{\mathcal{F}_i\}_{i=1,2,\dots C}$ with the same neural topology consisting of three
convolutional layers with batch normalization (BN) and ReLU activations, i.e.,
$\mathbf{y}=\mathbf{x}+\sum_{i=1}^C\mathcal{F}_i(\mathbf{x})$.
The first and third convolution layers in each branch use $1{\times}1$ kernels and the
second one uses $3{\times}3$ kernels. The first one embeds input features of
width $W$ to small-sized bottleneck features of width $B$, and the third one embeds reversely.
Namely, {the width of both input and output} of the second convolutional layer is $B$.
See Figure \ref{fig:iresnext-3} illustrating a block of ResNeXt architecture with $C=9$.
We often say $C$ as the number of branches or cardinality. Note that each branch
$\mathcal{F}_i$ can be parameterized by convolutional weights $\mathbf{w}_i$ and BN
parameters $\mathbf{u}_i$, thus one can write the block as
\begin{equation*}\label{eq:resnext}
\mathbf{y}=\mathbf{x}+\mathcal{F}(\mathbf{x};\mathbf{w}_1,\mathbf{u}_1)+\cdots+\mathcal{F}(\mathbf{x};\mathbf{w}_C,\mathbf{u}_C).
\end{equation*}
% Each branch $\mathcal{T}_i(\mathbf{x})=\mathcal{T}(\mathbf{x};\mathbf{w}_i,\mathbf{u}_i)$ can be parameterized by convolutional weights $\mathbf{w}_i$ and BN parameters $\mathbf{u}_i$.
% Then, one can write the $l$-th module of ResNeXt as %can be formally written as
% \begin{equation*}\label{eq:resnext}
% \mathbf{y}^{(l)}=\mathbf{x}^{(l)}+\sum_{i=1}^C\mathcal{T}(\mathbf{x}^{(l)};\mathbf{w}_i^{(l)},\mathbf{u}_i^{(l)}),
% \end{equation*}
% where $(\mathbf{x}^{(l)},\mathbf{y}^{(l)})$ be a pair of input and output vectors of the $l$-th module.

In this paper, we primarily focus on ResNeXt for demonstrating our approaches, but in
principle, they are applicable for generic multi-branch, even non-convolutional
architectures. Moreover, it is often possible to understand many neural networks as
multi-branch ones even if they do not enforce them explicitly. For example, 2-layer
neural network $\mathbf{y}=V\sigma(U\mathbf{x})$  with an activation function $\sigma$
can be understood as a multi-branch architecture by letting
$\mathcal{F}_i(x)=V_{\cdot,i}\sigma(U_{i,\cdot}\mathbf{x})$, i.e., the number of
branches is that of columns in the weight matrix $V$. This implies that
ResNet with basic-type residual blocks and DenseNet with bottleneck-type blocks are
also multi-branch architectures since their blocks consist of two convolutional layers.

% We explain ResNet \cite{resnet_he2016} and ResNeXt \cite{resnext_xie2017} in this subsection since our
% methods and architectures are based on them. ResNet consists of residual blocks which can be expressed as
% $\mathbf{y}=\mathbf{x}+\mathcal{F}(\mathbf{x})$ where $\mathcal{F}$ is a nonlinear residual function.
% The function of a bottleneck-type block consists of three convolutional layers with batch
% normalization (BN) \cite{bn_ioffe2015} and ReLU. The first and third convolution layers use $1\times1$ kernels
% and the second one uses $3\times3$ kernels. In general, the first one embeds input features of width $W$ to
% small-sized bottleneck features of width $B$, and the third one embeds reversely. Namely, the second convolutional
% layer uses $B$ as width of input and output features.

% ResNeXt \cite{resnext_xie2017}, an extension of ResNet, uses multiple residual functions which have same topology,
% i.e., $\mathbf{y}=\mathbf{x}+\mathcal{F}_1(\mathbf{x})+\cdots+\mathcal{F}_C(\mathbf{x})$
% where $C$ be the number of residual functions or branches. Note that $C$ is called as cardinality \cite{resnext_xie2017}.
% One can observe that the width of whole bottleneck features becomes $B\times C$.
% Increasing cardinality is able to improve network
% performance as well as increasing width $W,\;B$ or depth. This residual block can be implemented efficiently by
% using grouped convolutions. The network illustrated in Figure \ref{fig:iresnext-3} is a ResNeXt of cardinality $C=9$.

\section{Anytime prediction with thin neural sub-networks}

As described in Section \ref{sec:pre:shallow}, training shallow sub-networks jointly is difficult and thus
the overall performance of the sub-networks is degraded.
% since shallow ones cannot
% capture both coarse-level and high-level features.
In this section %, on contrary to this,
we study another direction maintaining the same depth of sub-networks, i.e., build thin sub-networks of ResNeXt or its variants. We also introduce a new anytime prediction task using a hierarchical taxonomy.
%, under minor
%modifications from a mutli-branch architecture, ResNeXt.

\subsection{Inclusive ResNeXt}\label{sec:method:iresnext}

\begin{figure}[t]
    \centering
    \begin{subfigure}{0.3\textwidth}
        \includegraphics[width=\textwidth]{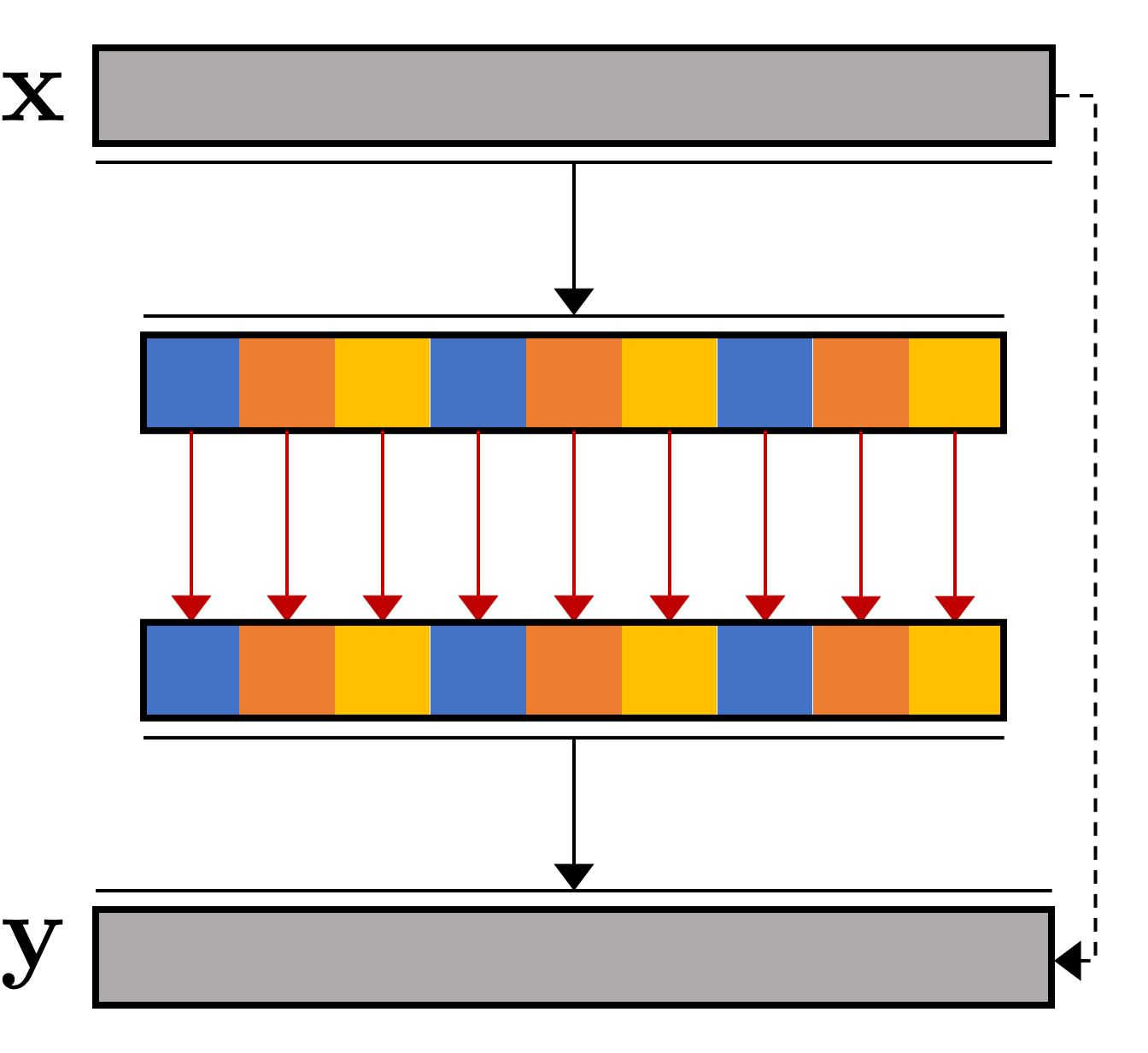}
        \caption{Full-network}
        \label{fig:iresnext-3}
    \end{subfigure}
    \hfill
    \begin{subfigure}{0.3\textwidth}
        \includegraphics[width=\textwidth]{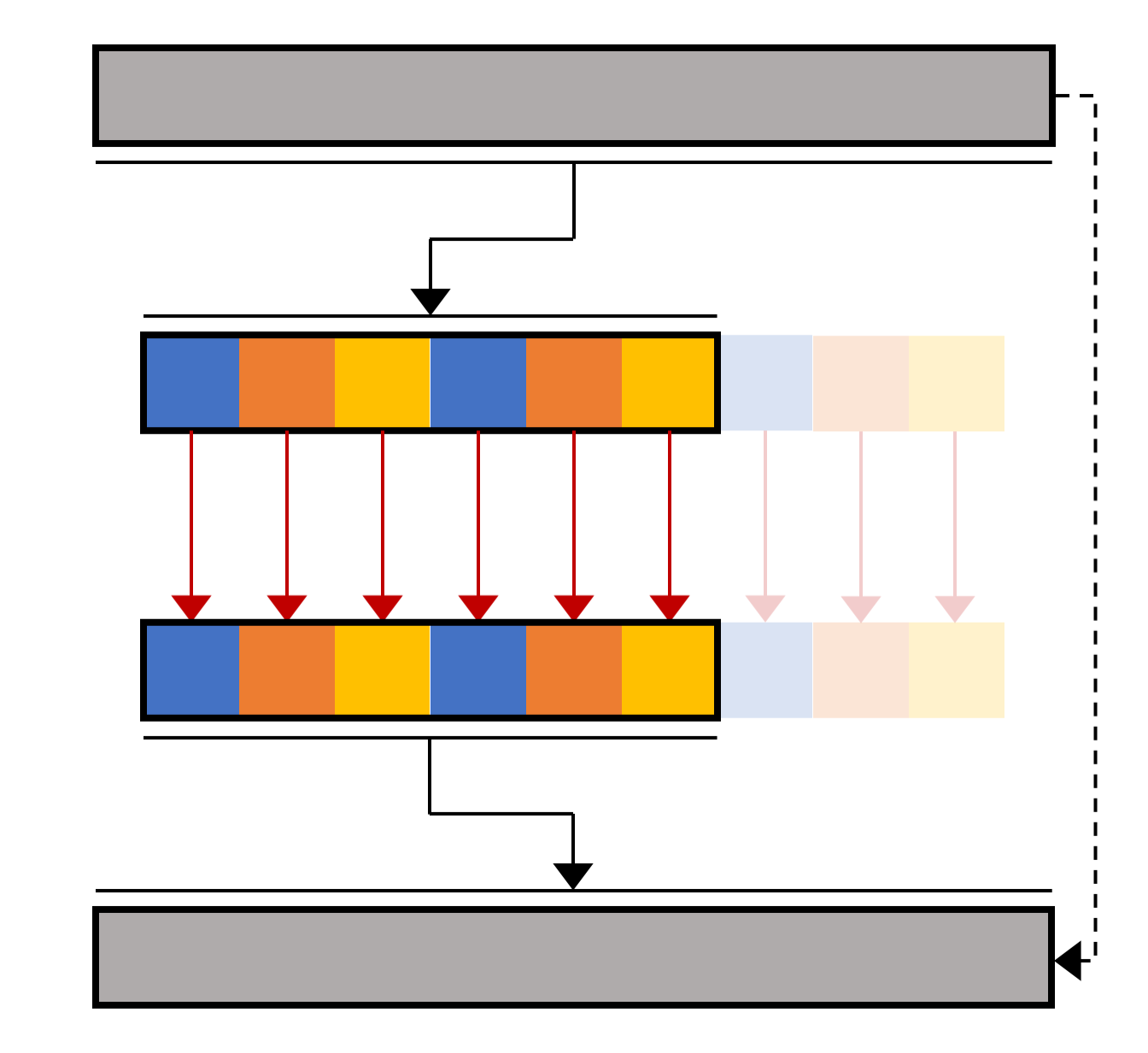}
        \caption{Wide sub-network}
        \label{fig:iresnext-2}
    \end{subfigure}
    \hfill
    \begin{subfigure}{0.3\textwidth}
        \includegraphics[width=\textwidth]{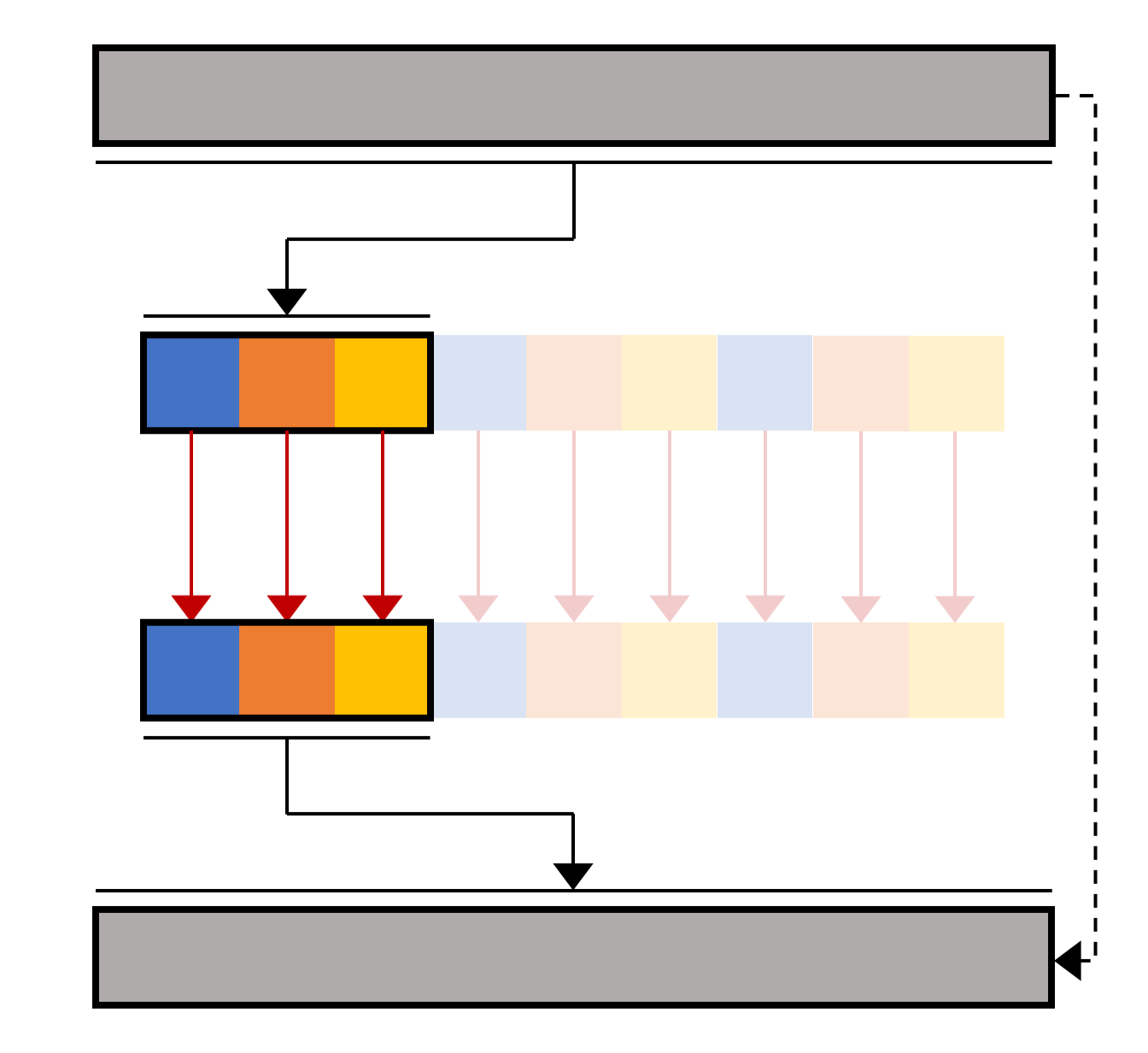}
        \caption{Thin sub-network}
        \label{fig:iresnext-1}
    \end{subfigure}
    \caption{I-ResNeXt. The black and red solid lines are convolutions of
    $1{\times}1$ and $3{\times}3$ kernels, respectively. The dashed
    lines are identity functions. Each sub-network is built by removing branches from ResNeXt architecture.
    Note that sub-networks are also ResNeXt and %their inference computation cannot be amortized across layers, i.e.,
    the $i$-th layer's inference result of a sub-network cannot be re-used for a different sub-network.
    %they have distinct outputs even if inputs are {\color{red}identical}.
    % This figure shows the
    % full-network (a) of $C=9$ and 2 sub-networks (b), (c)
    % by removing 3 and 6 branches from (a), respectively. In this case, the number of sub-networks is $K=3$.
    This figure shows (a) the full-network and (b), (c) sub-networks of I-ResNeXt of $C=9,\;K=3$.}
    \label{fig:iresnext}
\end{figure}

To build thin sub-networks from ResNeXt, we just remove the same number of branches for each block.
It means that each block is expressed by
$\mathbf{y}=\mathbf{x}+\mathcal{F}_1(\mathbf{x})+\cdots+\mathcal{F}_{C'}(\mathbf{x})$
where $C'\le C$ is the number of remaining branches. As the cardinality decreases, the width of
bottleneck in residual blocks also decreases, i.e., smaller $C'$ implies thinner sub-networks.
For simplicity, we use $C_k=kC/K$ as cardinality of {the} $k$-th sub-network where $K$ be the number of
sub-networks of interest. Note that the maximum number of sub-networks is $K=C$. Since the $(k+1)$-th sub-network includes
the $k$-th one, we refer this as Inclusive ResNeXt (I-ResNeXt) that is % These sub-networks
illustrated in Figure \ref{fig:iresnext}.
To produce final outputs for classification, we use an independent, auxiliary classifier $g_k$ for each
sub-network. {We choose a shallow classifier consisting of a batch normalization, a ReLU, a global average pooling, and
a fully connected layers sequentially, like ResNet-based architectures. Hence, the total number of the auxiliary parameters is quite negligible
compared to that of the original ones.}
Then, we can express a sub-network $f_k$ in \eqref{eq:anytime-loss} as follows: %the following:
\begin{equation*}
    f_k(\mathbf{x})=g_k(\mathbf{x}^{(L+1)}_k),
    \quad \mathbf{x}^{(l+1)}_k=\mathbf{x}^{(l)}_k+\mathcal{F}(\mathbf{x}^{(l)}_k;\mathbf{w}_1^{(l)},\mathbf{u}_1^{(l)})+
    \cdots+\mathcal{F}(\mathbf{x}^{(l)}_k;\mathbf{w}_{C_k}^{(l)},\mathbf{u}_{C_k}^{(l)}),
    \quad \mathbf{x}^{(1)}_k=\mathbf{x}
\end{equation*}
where $L$ is the number of blocks. All sub-networks can be trained jointly by optimizing \eqref{eq:anytime-loss}.

% After joint-training, as shown in Figure \ref{fig:resnext-norm}, the $\ell_1$-norm of weight of the third
% convolutional layer of a branch in the fifth block of 29-layer I-ResNeXt decreases as the index of the branch
% increases, while the original ResNeXt does not. This implies that I-ResNeXt utilizes the parameters of
% sub-networks in a progressive and efficient manner.

\textbf{Independent batch normalization (BN) layers.}
During training neural networks, the input distribution at some layer  changes as the previous layers
are updated. This makes training parameters be more difficult. The batch normalization \cite{bn_ioffe2015}
layer alleviates this phenomenon by normalizing the input distribution as zero-mean and unit-variance.
However, in our case, the number of branches is also changed while optimizing the anytime loss
\eqref{eq:anytime-loss}. Thus, the distribution of $\mathbf{x}_k^{(l)}$ of the $k$-th sub-network can
be also changed depending on $k$. Therefore, normalizing the input distribution by universal (or shared)
BN parameters $\mathbf{u}_i$ regardless of $k$ may not work.
To handle this issue, we use independent
BN parameters $\mathbf{u}_{i,k}$ for each sub-network, i.e.,
\begin{equation*}\label{eq:indep_bn}
\mathbf{x}_k^{(l+1)}=
\mathbf{x}_k^{(l)}+
%\sum_{i=1}^{C_k}\mathcal{F}(\mathbf{x}_k^{(l)};\mathbf{w}_i^{(l)},\mathbf{u}_{i,k}^{(l)}).
\mathcal{F}(\mathbf{x}_k^{(l)};\mathbf{w}_1^{(l)},\mathbf{u}_{1,k}^{(l)})+\cdots+
\mathcal{F}(\mathbf{x}_k^{(l)};\mathbf{w}_{C_k}^{(l)},\mathbf{u}_{C_k,k}^{(l)}).
\end{equation*}
We emphasize that the number of newly introduced parameters that are not {shared} among sub-networks
is quite negligible compared to other shared parameters, where they are highly effective for obtaining
high-performance sub-networks under the anytime loss \eqref{eq:anytime-loss}. We provide experimental
supports of this effect in Section \ref{sec:res:ablation}.

\subsection{Inclusive Sparse ResNeXt}\label{sec:method:isresnext}

\begin{figure}[t]
    \centering
    \begin{subfigure}{0.3\textwidth}
        \includegraphics[width=\textwidth]{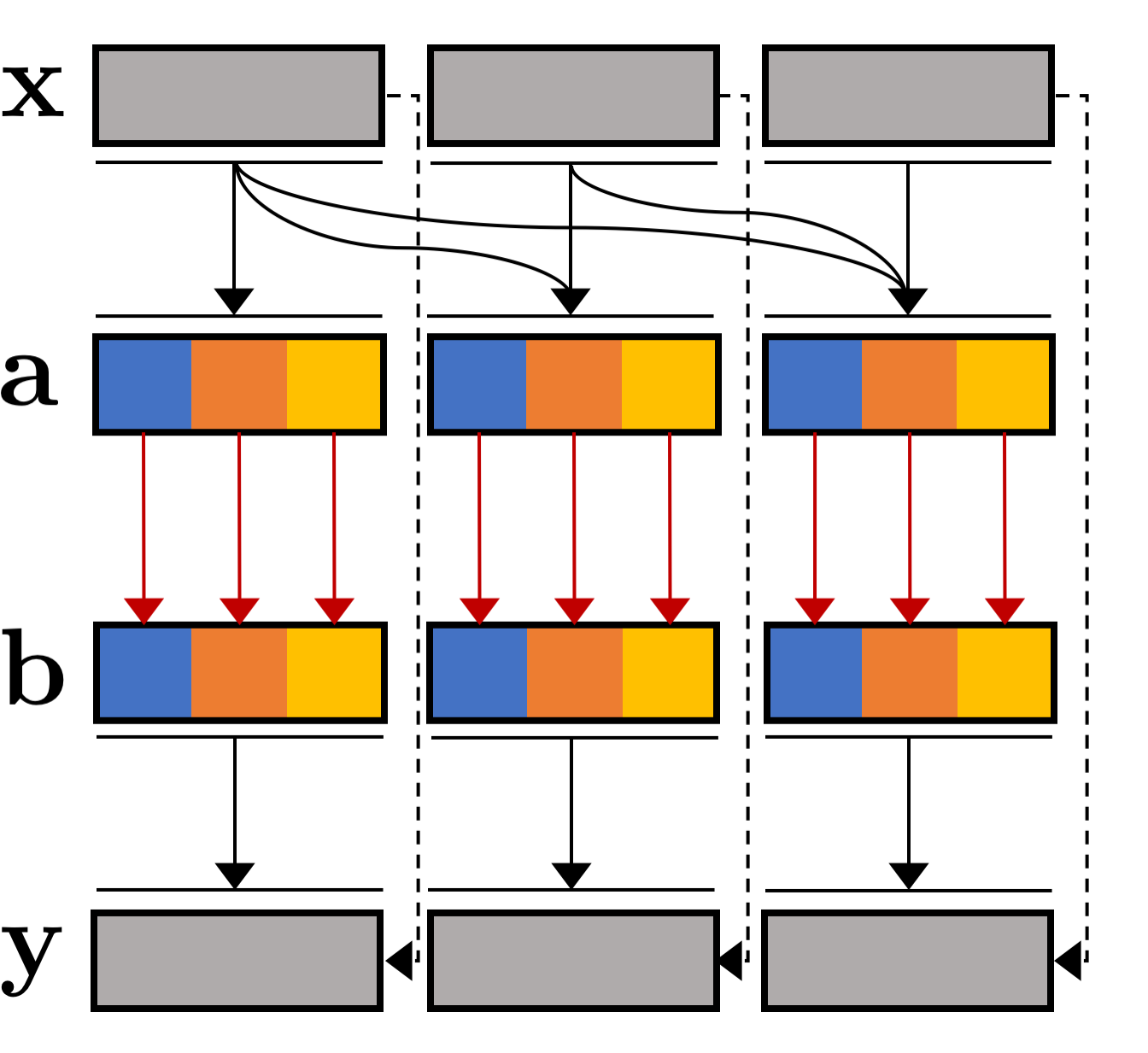}
        \caption{Full-network}
        \label{fig:isresnext-3}
    \end{subfigure}
    \hfill
    \begin{subfigure}{0.3\textwidth}
        \includegraphics[width=\textwidth]{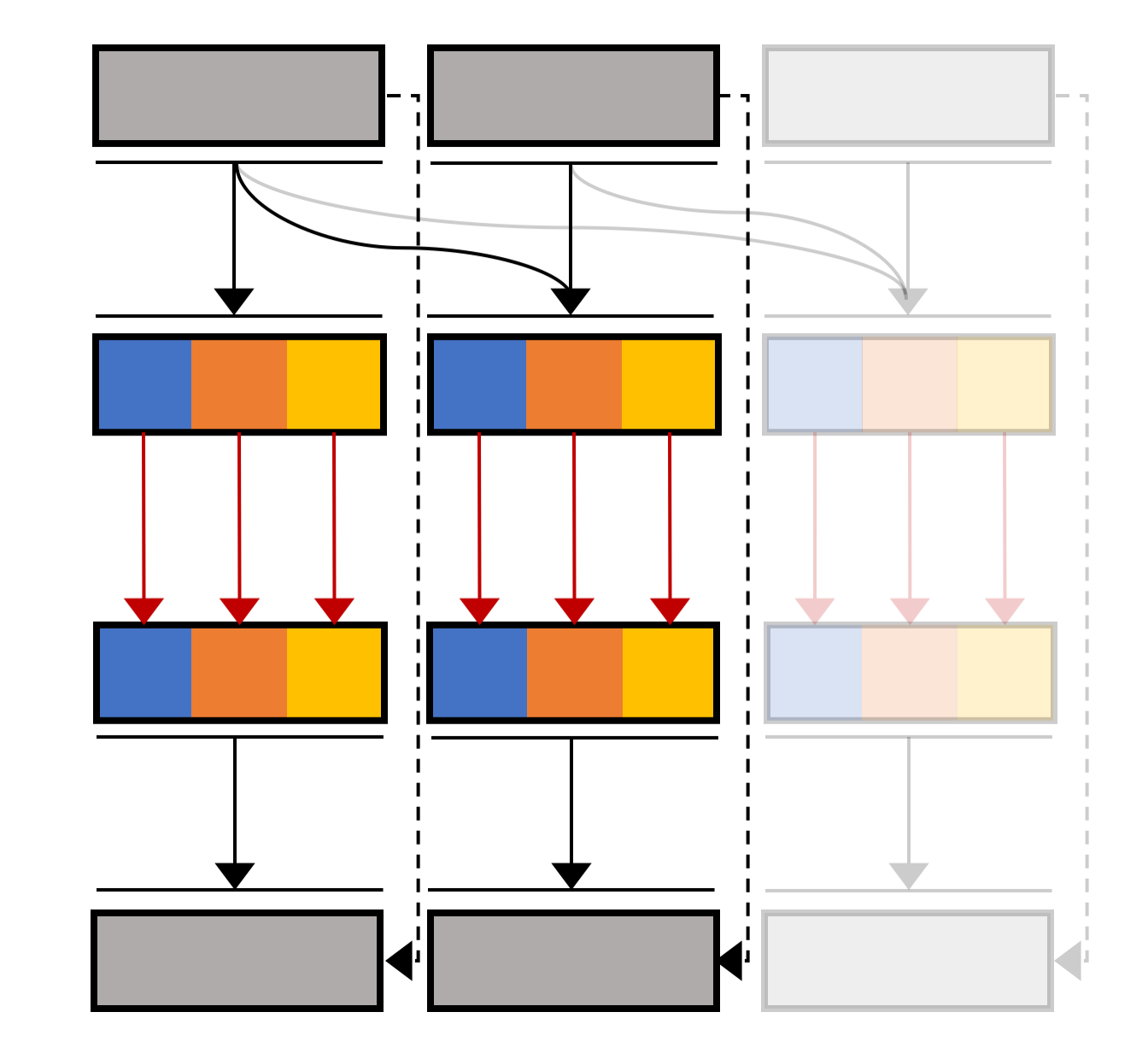}
        \caption{Wide sub-network}
        \label{fig:isresnext-2}
    \end{subfigure}
    \hfill
    \begin{subfigure}{0.3\textwidth}
        \includegraphics[width=\textwidth]{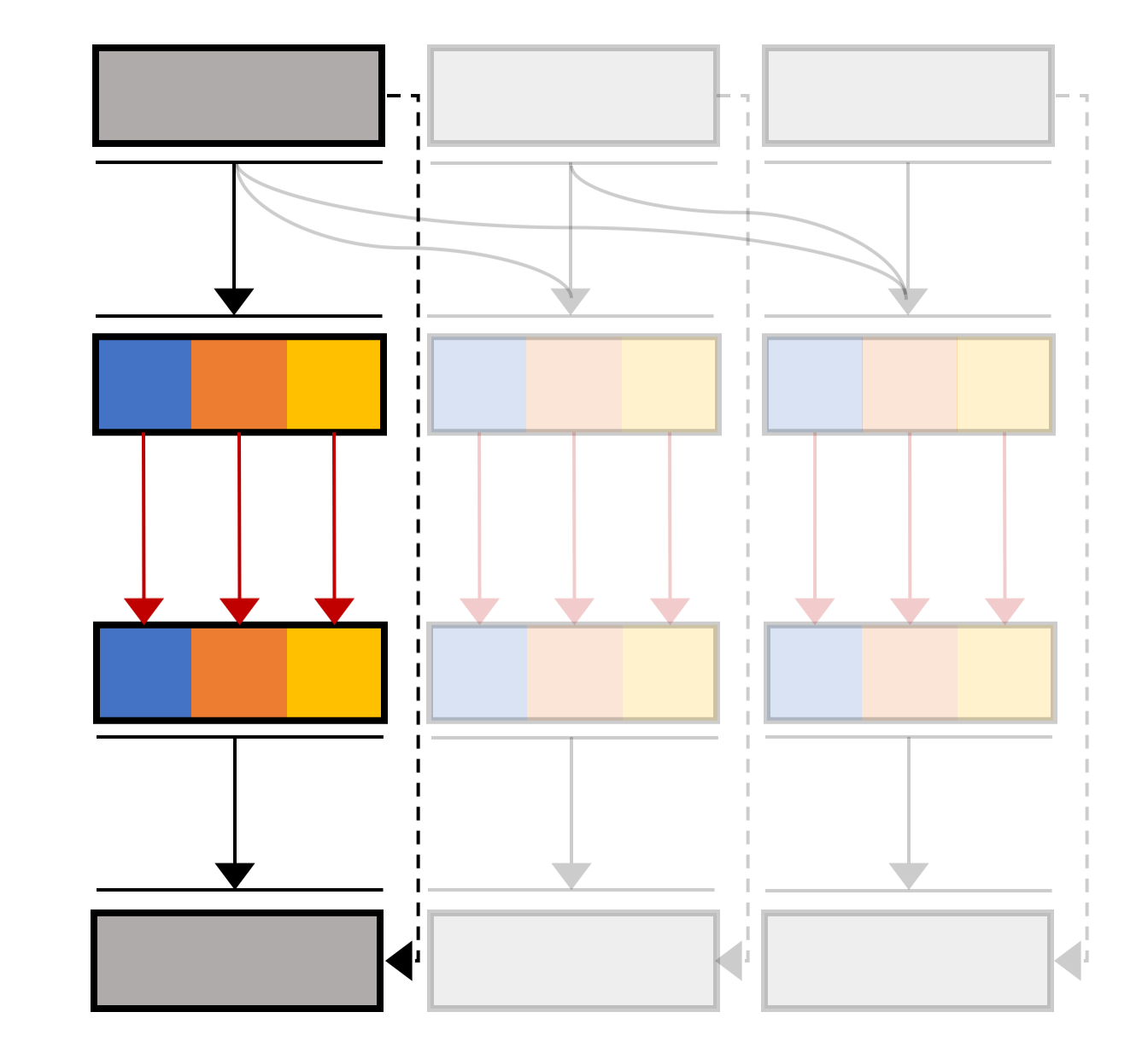}
        \caption{Thin sub-network}
        \label{fig:isresnext-1}
    \end{subfigure}
    \caption{IS-ResNeXt. Each sub-network can be built by removing branches and reducing width of features from IS-ResNeXt
    architecture as described in Section \ref{sec:method:isresnext}. Similarly to I-ResNeXt, each sub-network is also
    IS-ResNeXt. Note that the $k$-th output sub-feature only depends on $1,\ldots,k$-th input sub-features. This property helps
    joint-training with thin sub-networks be efficient.
    This figure shows (a) the full-network and (b), (c) sub-networks of IS-ResNeXt of $C=9,\;K=3$.}
    \label{fig:isresnext}
\end{figure}

The main issue of I-ResNext is the increased number of propagations in its training.
Since the intermediate features are different among sub-networks, K propagation steps
are required for training $K$ sub-networks of I-ResNeXt at each iteration, i.e.,
the overall training time is
$K$ times slower than training a single ResNeXt. Due to the same reason, I-ResNext does not have the interruptibility of anytime prediction.
%Since training some DNNs on large-scale datasets such as ImageNet often takes couple of weeks,
%it is hard to train I-ResNeXt on such datasets. 
To resolve the issue, we propose a new architecture, called 
Inclusive Sparse ResNeXt (IS-ResNeXt), by enforcing a certain sparsity on I-ResNeXt.

First, we split the
{input feature $\mathbf{x}$ and the output feature $\mathbf{y}$ of a block} in {channel-wise} into $K$ sub-features 
$\mathbf{x}=[\mathbf{x}_1;\ldots;\mathbf{x}_K]$,
$\mathbf{y}=[\mathbf{y}_1;\ldots;\mathbf{y}_K]$. 
Our goal is to make the $k$-th output sub-feature $\mathbf{y}_k$ be computable using % only depend on
only $\mathbf{x}_1,\ldots,\mathbf{x}_k$, irrespectively of  
%, then we can obtain consistent sub-features $\mathbf{y}_k$
whether $\mathbf{x}_{k+1},\ldots,\mathbf{x}_K$ is removed or not. 
Then, by using $[\mathbf{y}_1;\ldots;\mathbf{y}_k]$ as
the output feature of the $k$-th sub-network, 
one can obtain output features of all sub-networks by only a single forward
propagation.

To describe the details, %our idea to achieve the goal,
let $\mathbf{a}_1,\ldots,\mathbf{a}_C$ and $\mathbf{b}_1,\ldots,\mathbf{b}_C$ be the first and second intermediate features
of a block in I-ResNeXt.
Remark again that $\mathbf{a}_i$ and $\mathbf{b}_i$ have width $B$, and they are
the input and output of $3{\times}3$ $i$-th convolution in the second layer, respectively. 
%Further
{In the case of I-ResNeXt, one can observe that
$\mathbf{a}_i$ is a function of
$\mathbf{x}_1,\ldots,\mathbf{x}_K$ and $\mathbf{y}_k$ is a function of $\mathbf{b}_1,\ldots,\mathbf{b}_C$,
i.e., $\{\mathbf{a}_i\}$ and $\{\mathbf{b}_i\}$ are densely connected to $\{\mathbf{x}_k\}$ and $\{\mathbf{y}_k\}$,
respectively.
}

As illustrated in Figure \ref{fig:isresnext}, IS-ResNeXt enforces
{%\color{red}
that
$\mathbf{a}_i$ be a function of
$\mathbf{x}_1,\ldots,\mathbf{x}_{\lceil iK/C\rceil}$ and $\mathbf{y}_k$ be a function of
$\mathbf{b}_{(k-1)C/K+1},\ldots,\mathbf{b}_{kC/K}$, i.e., %to obtain the consistent features. In detail,
}
\begin{align*}
\mathbf{a}_i & =\text{conv}_{1{\times}1}(\text{ReLU}(\text{BN}([\mathbf{x}_1;\ldots;\mathbf{x}_{\lceil iK/C\rceil}]))), \\
\mathbf{b}_i & =\text{conv}_{3{\times}3}(\text{ReLU}(\text{BN}(\mathbf{a}_i))), \\
\mathbf{y}_k & =\text{conv}_{1{\times}1}(\text{ReLU}(\text{BN}([\mathbf{b}_{(k-1)C/K+1};\ldots;\mathbf{b}_{kC/K}])))+\mathbf{x_k}.
\end{align*}
{
Namely, $\{\mathbf{a}_i\}$ and $\{\mathbf{b}_i\}$ are sparsely connected to $\{\mathbf{x}_k\}$ and $\{\mathbf{y}_k\}$,
respectively. This sparsity provides the interruptible property, i.e., one can compute $\mathbf{y}_k$
sequentially.
%only depends on $\mathbf{x}_1,\ldots,\mathbf{x}_k$.
}
%Note that the above equations can be simply written by $\mathbf{y}=\mathbf{x}+\mathcal{F}(\mathbf{x})$.
Similarly to I-ResNeXt, we have the following restricted model $f_k$:
\begin{equation*}
    f_k(\mathbf{x})=g_k([\mathbf{x}^{(L+1)}_1;\ldots;\mathbf{x}^{(L+1)}_k]),
    \quad \mathbf{x}^{(l+1)}=\mathbf{x}^{(l)}+\mathcal{F}(\mathbf{x}^{(l)};\mathbf{w}^{(l)},\mathbf{u}^{(l)}),
    \quad \mathbf{x}^{(1)}=\mathbf{x}
\end{equation*}
where $\mathbf{w}$ and $\mathbf{u}$ are parameters for convolutional and BN layers, respectively.
The convolutions in the third layer can be implemented by one grouped convolution like the second layer.
%Since each sub-feature does not depend on sub-networks, we use shared BN layers for all sub-networks.
%Note that BN layers must be shared so that the sub-features are not dependent on sub-networks.
We note that BN layers must be shared in IS-ResNeXt (in contrast to I-ResNeXt) since its sub-features are shared (or reused) among different sub-networks.
{We} emphasize again that one can obtain all outputs of all sub-networks of IS-ResNeXt 
by only a single forward propagation, and also %we can
{compute} the gradient of \eqref{eq:anytime-loss} by only a single backward propagation.

\subsection{Hierarchical anytime prediction}
Under the anytime prediction task, a smaller sub-network inevitably provides a lower accuracy that
might not match usable accuracy in practical applications.
%we show that our proposed anytime architectures provide significantly better accuracy, in particular under small FLOPs,
%compared to the state-of-art model.
%However, it is still much lower than that using large FLOPs.
To address the issue alternatively, we reformulate
the anytime prediction task using 
a hierarchical taxonomy,
where the taxonomy can be 
% either included in a dataset (e.g., CIFAR-100~\cite{krizhevsky2009learning}) or 
extracted from the natural language information, e.g., WordNet \cite{wordnet_miller1995}.
%Here note that it has been demonstrated that there is a strong correlation between hierarchical semantic relationships and the visual appearance of objects~\cite{deng2010does}.
This approach is also motivated by a strong empirical correlation between hierarchical semantic relationships and the visual appearance of objects~\cite{deng2010does}.
In the proposed
hierarchical anytime prediction task,
a model can predict coarse labels when 
small budgets are allowed, otherwise predict the original fine labels.
%for maintaining performance anytime.
Formally, for an input $\mathbf{x}$, let {$\{y_{d}\}_{d=1,\ldots,D}$}
be its labels from fine to coarse, e.g., $\{\text{Afghan hound},\text{hound},\text{hunting dog},\text{dog}\}$
or $\{\text{albatross},\text{pelagic bird},\text{seabird},\text{aquatic bird},\text{bird}\}$.
To classify these labels, we add auxiliary classifiers to all sub-networks of IS-ResNeXt (or I-ResNeXt), i.e.,
\begin{equation*}
f_{k,d}(\mathbf{x})=g_{k,d}([\mathbf{x}_1^{(L+1)};\ldots;\mathbf{x}_k^{(L+1)}]),
\end{equation*}
where $g_{k,d}$ is a classifier attached to the $k$-th sub-network for predicting $y_d$.
For a given dataset
% Under an assumption that
% all images have same length (or depth of taxonomy) of labels, we can write a dataset as
%$\mathcal{D}=\{(\mathbf{x}_i,y_{i,1},\ldots,y_{i,D}):i=1,\ldots,N\}$,
$\mathcal{D}$,
one can define the following loss to train all sub-network jointly of multi-classifiers:
%assigning all-level labels to all sub-networks:
\begin{equation}\label{eq:hanytime-loss}
\begin{split}
% \mathcal{L}_\text{h-anytime}(\mathcal{D},f)=
% \frac{1}{N}\sum_{i=1}^N\sum_{k=1}^K\sum_{d=1}^D\mathcal{L}(y_{i,d},f_{k,d}(\mathbf{x}_i)),
\mathcal{L}_\text{h-anytime}(\mathcal{D},f)=
\frac{1}{|\mathcal{D}|}\sum_{(\mathbf{x},\{y_d\})\in\mathcal{D}}\sum_{k=1}^K\sum_{d=1}^D\mathcal{L}(y_{d},f_{k,d}(\mathbf{x})),
\end{split}
\end{equation}
which generalizes the original anytime loss \eqref{eq:anytime-loss}.
Once a model is trained by the above loss,
one can choose appropriate coarse and fine
labels targeted by small and large sub-networks
so that all of them can match a certain level of accuracy, e.g.,
that of the solely trained one (see Section \ref{sec:res:hanytime}).

% !TEX root = main.tex

\section{Experimental results}\label{sec:exp}

We evaluate our architectures for anytime prediction on CIFAR-10/100 \cite{cifar_krizhevsky2009a},
ImageNet \cite{imagenet_deng2009} and Caltech-UCSD Birds (CUB) \cite{cub_wah2011} datasets.
%, and Stanford Dogs \cite{dog_khosla2011}.
In this section, we denote the width of features as $W$, the number of sub-networks as $K$, the number of blocks as $L$.
We use different hyperparameters for $W,K,L$ {depending} on datasets and tasks.
The {details of} training setups and architectures are described in the supplementary material.
%Appendix \ref{sec:appendix:setup}.
In Section \ref{sec:res:ablation}, we verify that using independent batch normalization layers
improve the overall performance and 
utilizing thin sub-networks are more effective than shallow ones.
Then, we compare ours with existing anytime prediction models in Section \ref{sec:res:anytime}.
Finally, we apply our architecture to a new task, named
hierarchical anytime prediction, in Section \ref{sec:res:hanytime}.
%and then apply our architectures to the problem.

\begin{figure}[t]
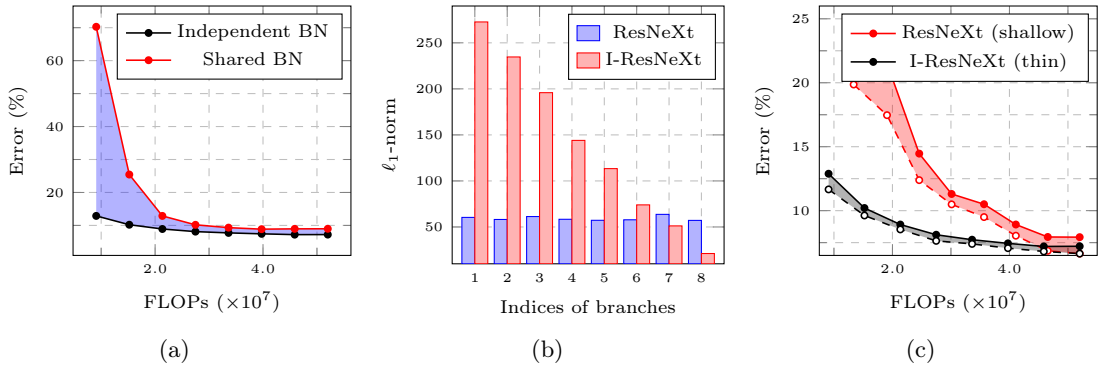

% \vspace{-2mm}
% \vspace{0.12in}
    \centering
    \begin{subfigure}{0.32\textwidth}
        \includegraphics[width=\textwidth,height=0.9\textwidth]{effect_bn}
        % \caption{Effect of independent BN}
        \caption{}
        \label{fig:effect-bn}
    \end{subfigure}
    \hfill
    \begin{subfigure}{0.32\textwidth}
        \includegraphics[width=\textwidth,height=0.9\textwidth]{weight_analysis}
        % \caption{$\ell_1$-norm of weights of branches.}
        \caption{}
        \label{fig:resnext-norm}
    \end{subfigure}
    \hfill
    \begin{subfigure}{0.32\textwidth}
        \includegraphics[width=\textwidth,height=0.9\textwidth]{comparison}
        % \caption{Shallow versus thin sub-networks}
        \caption{}
        \label{fig:comparison}
    \end{subfigure}
    \caption{
    (a) Classification errors of thin sub-networks which are trained jointly with independent
    or shared BN parameters. Using independent BN layers for sub-networks can improve the overall performance.
    (b) $\ell_1$-norm of convolutional weights of branches. In the original ResNeXt, all branches equally
    contribute to construct output features. On the other hand, the contribution of branches of I-ResNeXt
    decreases as the index of the branch increases.
    (c) Shallow versus thin sub-networks for anytime prediction.
    % The red and black lines are results of
    % using shallow and thin sub-networks, respectively, and the solid and dashed lines are results of training jointly
    % and independently, respectively.
    The jointly trained thin sub-networks outperform shallow ones under any FLOPs.
    Moreover, the joint-training with thin sub-networks does not hurt their performance compared
    to solely trained ones.}
\end{figure}

\subsection{Ablation study}\label{sec:res:ablation}

\textbf{Independent batch normalization layers.}
We first evaluate the effect of using independent BN layers
when all $K=8$ thin sub-networks of 29-layer I-ResNeXt
% of width $W=64$
are
jointly trained under the CIFAR-10 dataset. As shown in Figure \ref{fig:effect-bn}, the sub-networks have
poor performance when using shared BN layers. In particular, the thinnest sub-network has
more than $70\%$ classification error. %, in other words, the joint-training fails. On the contrary to this,
On the other hand, using independent BN layers makes all sub-networks
be trained stably, and reduces
the classification errors of all sub-networks.
As a result, the thinnest and full network achieve $12.9\%$ and $7.2\%$ errors which are
$81.7\%$ and $19.6\%$ smaller than those using shared BN layers, respectively.
% although few number of parameters are newly introduced.

After joint-training with independent BN layers, we observe that the branches of I-ResNeXt are learned
in a progressive manner. For example,
as shown in Figure \ref{fig:resnext-norm}, the $\ell_1$-norm of weight of the third
convolutional layer of a branch in the fifth block of the 29-layer I-ResNeXt decreases as the index of the branch
increases, while the original ResNeXt does not.
This implies that the branches modify features progressively to obtain better results when more branches are used.

\textbf{Shallow versus thin sub-networks.}
To verify that using thin sub-networks is more effective for anytime prediction than shallow ones,
we train shallow/thin sub-networks of ResNeXt jointly/separately and compare them.
To build shallow sub-networks from a 29-layer (i.e., 9 blocks) ResNeXt, % of width $W=64$,
we attach $K=8$ auxiliary classifiers to outputs of all blocks except the first one.
Each classifier is of the same architecture with that used for I-ResNeXt.
We use I-ResNeXt described in Section \ref{sec:method:iresnext} to build $K=8$ thin sub-networks.

We first train each sub-network separately on the CIFAR-10 dataset. The red and black dashed lines in Figure \ref{fig:comparison} are
classification errors of the solely-trained shallow and thin sub-networks, respectively. As shown in the figure, 
one can observe that
{the errors of shallow networks increase rapidly}
%shallow networks increase the error rapidly 
as the FLOPs, i.e., the depths, decrease, e.g., the shallowest network have
$19.9\%$ error.
However,
{the errors of thin networks drop less rapidly}
%thin networks drop the accuracy less rapidly
as
%all of them
{all of them}
have 29 layers, i.e., they can capture both coarse-level and high-level features.
In particular, the difference between errors of shallow and thin networks increases up to $8.2\%$ as the required FLOPs decrease.
This confirms that thin sub-networks are more effective for anytime prediction.
%if joint-training does not hurt their performance.

Next, we train all shallow (or thin) sub-networks
jointly under minimizing the anytime loss \eqref{eq:anytime-loss}, which is
to evaluate the performance drop caused by the joint-training. 
The
red and black solid lines in Figure \ref{fig:comparison} are the results of the joint-training with shallow and
thin sub-networks, respectively. As shown in the figure, we observe that joint-training of thin sub-networks
does not hurt their performance: the sub-networks
lose their classification accuracy at most $1.2\%$. However, jointly trained shallow sub-networks lose the accuracy
up to $5.1\%$, i.e., $4\sim 5$ times more.
Consequently, jointly trained thin sub-networks, i.e., I-ResNeXt, outperform shallow ones significantly under any FLOPs.
Remark that the thinnest one of I-ResNeXt has $48.4\%$ smaller error compared to 
the shallowest one even though the former
one has smaller FLOPs.
%From these results, one can conclude that joint training with thin sub-networks of the same depth does not cause
%the performance drop, and it means that thin sub-networks are more appropriate for anytime prediction.

% These results show that it is hard to capture coarse-level features in shallow sub-networks and
% joint-training with them is difficult

% additional classifiers to output features of blocks except the first one.  We train these sub-networks jointly by using \eqref{eq:anytime-loss} on
% CIFAR-10 dataset. Also we train each sub-network independently to measure loss of accuracy caused by joint-training.
% The result is shown in Figure \ref{fig:comparison}. One can observe
% that thin sub-networks have higher accuracy than shallow ones under any FLOPs. Especially, under small FLOPs,
% the thinnest sub-network has $x\%$ higher performance compared to the shallowest sub-network.
% Remark that efficiency on small FLOPs is useful on resource-limited environments such as mobile devices.
% Also, surprisingly, the joint-training with thin sub-networks does not hurt the overall performance.
% However, jointly trained sub-networks have poor performance compared to solely trained networks.
% These results justify using thin sub-networks for anytime prediction instead of shallow ones.

\subsection{Results for anytime prediction}\label{sec:res:anytime}

\begin{figure}[t]
\centering
\begin{tikzpicture}
    \begin{groupplot}[
        group style={
            group name=anytime,
            group size=3 by 1,
            xlabels at=edge bottom,
            ylabels at=edge left,
        },
        height=5.2cm,width=5.2cm,
        ymajorgrids=true,yminorgrids=true,
        xmajorgrids=true,xminorgrids=true,
        minor tick num=1,
        grid style=dashed,
        yticklabel style={/pgf/number format/fixed,
                          /pgf/number format/fixed zerofill,
                          /pgf/number format/precision=0},
        xticklabel style={/pgf/number format/fixed,
                          /pgf/number format/fixed zerofill,
                          /pgf/number format/precision=1},
        tick label style = {font = \tiny},
        label style = {font = \scriptsize},
        legend style = {font = \scriptsize},
        ylabel=Error($\%$),
    ]
    \nextgroupplot[mark size=1.4pt,xlabel={FLOPs ($\times10^8$)},ymax=13,xmax=1.35] % CIFAR-10
%    \node[text width=5.2cm,align=center,anchor=north] at ([yshift=-5mm]my plots c1r1.south) {\captionof{subfigure}{CIFAR-10\label{fig:anytime-result:cifar10}}};
        % \node {\subcaption{asdf}};
        %\addlegendentry{MSDNet}
        \addplot[mark=triangle*,solid,line width=0.6pt,color=red,mark size=1.6pt]
        coordinates {
        (0.23638218,12.092)
        (0.33367722, 9.922)
        (0.44571786, 8.952)
        (0.63292082, 7.618)
        (0.71018642, 7.094)
        (0.79924850, 6.942)
        (0.90010706, 6.886)
        (1.02148866, 6.728)
        (1.07491554, 6.748)
        (1.13424066, 6.734)
        (1.19946402, 6.736)
        };\label{plots:msdnet}
        \coordinate (top) at (rel axis cs:0,1);
        
        %\addlegendentry{I-ResNeXt}
        \addplot[mark=*,solid,line width=0.6pt,color=black]
        coordinates {
        (0.18635274, 9.788)
        (0.32942602, 7.714)
        (0.47249930, 6.748)
        (0.61557258, 6.382)
        (0.75864586, 6.164)
        (0.90171914, 6.104)
        (1.04479242, 5.942)
        (1.18786570, 6.014)
        };\label{plots:iresnext}
        
        %\addlegendentry{IS-ResNeXt}
        \addplot[mark=*,solid,line width=0.6pt,color=blue,mark options={solid,fill=white}]
        coordinates {
        (0.13873802, 10.018)
        (0.30696714,  7.822)
        (0.50468746,  6.922)
        (0.73189898,  6.442)
        (0.98860170,  6.260)
        (1.27479562,  6.196)
        };\label{plots:isresnext}
        
        %\addlegendentry{DenseNet-BC}
        \addplot[mark=+,only marks,color=red,mark size=2pt,line width=0.8pt]
        coordinates {
        (0.74459698,  7.264)
        (1.28013928,  6.046)
        };\label{plots:densenet}
        
        % \addlegendentry{ResNeXt}
        \addplot[mark=x,only marks,color=black,mark size=2pt, line width=0.8pt]
        coordinates {
        (0.52005386,  6.518)
        (1.18786570,  5.564)
        };\label{plots:resnext}
        
        % \addlegendentry{ResNet}
        \addplot[mark=diamond,only marks,color=black,line width=0.8pt,mark size=2pt]
        coordinates {
        (0.71975562,  6.934)
        (1.29172106,  6.190)
        };\label{plots:resnet}
        
        % % NestedNet
        % \addplot[mark=square,solid,line width=0.6pt,color=cyan]
        % coordinates {
        % %(0.14779018, 32.438)
        % %(0.29078154, 20.520)
        % (0.57869578, 11.340)
        % (1.14779402,  8.946)
        % };\label{plots:nestednet}
        % \addplot[mark=square,dashed,line width=0.6pt,color=cyan]
        % coordinates {
        % (0.14779018, 32.438)
        % (0.29078154, 20.520)
        % (0.57869578, 11.340)
        % %(1.14779402,  8.946)
        % };
        
        % % NOT COMPLETED
        % \addplot[mark=square,solid,line width=0.6pt,color=cyan]
        % coordinates {
        % %(0.13181578, 58.40)
        % (0.37373194, 11.60)
        % (0.72574858,  6.86)
        % (1.18786570,  6.92)
        % };
        % \addplot[mark=square,dashed,line width=0.6pt,color=cyan]
        % coordinates {
        % (0.13181578, 58.40)
        % (0.37373194, 11.60)
        % %(0.72574858,  6.86)
        % %(1.18786570,  6.92)
        % };

        % % NestedNet (reported in paper)
        % \addplot[mark=square,solid,line width=0.6pt,color=cyan]
        % coordinates {
        % ( 0.29078154, 9.713483146)
        % ( 0.71975562, 8.011235955)
        % ( 4.56051210, 5.792134831)
        % (11.37248778, 5.258426966)
        % };\label{plots:nestednet}
    \nextgroupplot[mark size=1.4pt,xlabel={FLOPs ($\times10^8$)},ymax=44,ytick={25,31,...,43},xmax=1.35] % CIFAR-100
        % \addlegendentry{MSDNet}
        \addplot[mark=triangle*,solid,line width=0.6pt,color=red,mark size=1.6pt]
        coordinates {
        (0.23658468, 42.316)
        (0.33392292, 36.694)
        (0.44600676, 33.680)
        (0.63309812, 29.722)
        (0.71040692, 28.560)
        (0.79951220, 27.722)
        (0.90041396, 27.348)
        (1.02167316, 26.622)
        (1.07514324, 26.538)
        (1.13451156, 26.446)
        (1.19977812, 26.462)
        };
        
        % \addlegendentry{I-ResNeXt}
        \addplot[mark=*,solid,line width=0.6pt,color=black]
        coordinates {
        (0.18658404, 33.934)
        (0.32965732, 29.180)
        (0.47273060, 27.220)
        (0.61580388, 26.214)
        (0.75887716, 25.548)
        (0.90195044, 25.284)
        (1.04502372, 24.942)
        (1.18809700, 24.916)
        };

        % \addlegendentry{IS-ResNeXt}
        \addplot[mark=*,solid,line width=0.6pt,color=blue,mark options={solid,fill=white}]
        coordinates {
        (0.13879652, 37.204)
        (0.30708324, 31.038)
        (0.50486116, 27.908)
        (0.73213028, 26.760)
        (0.98889060, 26.242)
        (1.27514212, 26.072)
        };
        
        %\addlegendentry{DenseNet-BC}
        \addplot[mark=+,only marks,color=red,mark size=2pt, line width=0.8pt]
        coordinates {
        (0.74471668, 29.604)
        (1.28031568, 26.760)
        };
        
        % \addlegendentry{ResNeXt}
        \addplot[mark=x,only marks,color=black,mark size=2pt,line width=0.8pt]
        coordinates {
        (0.52028516, 26.898)
        (1.18809700, 24.048)
        };
        
        % \addlegendentry{ResNet}
        \addplot[mark=diamond,only marks,color=black,line width=0.8pt,mark size=2pt]
        coordinates {
        (0.71981412, 29.992)
        (1.29177956, 28.062)
        };
        
        % % NestedNet
        % \addplot[mark=square,solid,line width=0.6pt,color=cyan]
        % coordinates {
        % %(0.14784868, 80.744)
        % %(0.29084004, 60.084)
        % (0.57881188, 43.896)
        % (1.14791012, 38.290)
        % };
        % \addplot[mark=square,dashed,line width=0.6pt,color=cyan]
        % coordinates {
        % (0.14784868, 80.744)
        % (0.29084004, 60.084)
        % (0.57881188, 43.896)
        % %(1.14791012, 38.290)
        % };

        % % NestedNet (NOT COMPLETED)
        % \addplot[mark=square,solid,line width=0.6pt,color=cyan]
        % coordinates {
        % %(0.13187428, 95.75)
        % (0.37384804, 34.52)
        % (0.72592228, 29.37)
        % (1.18809700, 30.82)
        % };
        % \addplot[mark=square,dashed,line width=0.6pt,color=cyan]
        % coordinates {
        % (0.13187428, 95.75)
        % (0.37384804, 34.52)
        % %(0.72592228, 29.37)
        % %(1.18809700, 30.82)
        % };

        % % NestedNet (reported in paper)
        % \addplot[mark=square,solid,line width=0.6pt,color=cyan]
        % coordinates {
        % ( 0.29084004, 35.7)
        % ( 0.71981412, 32.9)
        % ( 4.56074340, 25.7)
        % (11.37271908, 24.5)
        % };
    \nextgroupplot[mark size=1.4pt,xlabel={FLOPs ($\times10^9$)}] % ImagetNet
        % \addlegendentry{MSDNet}
        \addplot[mark=triangle*,solid,line width=0.6pt,color=red,mark size=1.6pt]
        coordinates {
        (0.635656712, 41.756)
        (0.972944904, 34.178)
        (1.395106136, 31.838)
        (1.708362624, 31.064)
        (1.760538240, 30.950)
        };
        
        % \addlegendentry{IS-ResNeXt}
        % \addplot[mark=*,solid,line width=0.6pt,color=blue,mark options={solid,fill=white}]
        % coordinates {
        % (0.184415464, 	45.233)
        % (0.407365096, 	36.370)
        % (0.668849896, 	33.065)
        % (0.968869864, 	31.943)
        % (1.307425000, 	31.665)
        % (1.684515304, 	31.610)
        % };
        
        \addplot[mark=*,solid,line width=0.6pt,color=blue,mark options={solid,fill=white}]
        coordinates {
        (0.215650024, 42.700)
        (0.479468008, 34.244)
        (0.791454952, 31.352)
        (1.151610856, 30.992)
        (1.559935720, 30.998)
        };

        \coordinate (bot) at (rel axis cs:1,0);
    \end{groupplot}
    % legend
    \path (top|-current bounding box.north)--
          coordinate(legendpos)
          (bot|-current bounding box.north);
    \matrix[
        matrix of nodes,
        font=\scriptsize,
        anchor=south,
        draw,
        inner sep=0.2em,
        draw
    ]at([yshift=1ex]legendpos)
    {
        \ref{plots:iresnext}  & I-ResNeXt   &[5pt]
        \ref{plots:isresnext} & IS-ResNeXt  &[5pt]
        \ref{plots:msdnet}    & MSDNet      &[5pt]
        % \ref{plots:nestednet} & NestedNet   &[5pt]
        \ref{plots:densenet}  & DenseNet    &[5pt]
        \ref{plots:resnext}   & ResNeXt     &[5pt]
        \ref{plots:resnet}    & ResNet      \\
    };
    \node [text width=5.2cm,anchor=north,align=center] at ([yshift=-7mm]anytime c1r1.south) {\subcaption{CIFAR-10\label{fig:anytime-result:cifar10}}};
    \node [text width=5.2cm,anchor=north,align=center] at ([yshift=-7mm]anytime c2r1.south) {\subcaption{CIFAR-100\label{fig:anytime-result:cifar100}}};
    \node [text width=5.2cm,anchor=north,align=center] at ([yshift=-7mm]anytime c3r1.south) {\subcaption{ImageNet\label{fig:anytime-result:imagenet}}};
    % \node[text width=5.2cm,align=center,anchor=north] at ([yshift=-5mm]my plots c1r1.south) {\captionof{subfigure}{CIFAR-10\label{fig:anytime-result:cifar10}}};
    % \node[text width=5.2cm,align=center,anchor=north] at ([yshift=-5mm]my plots c2r1.south) {\captionof{subfigure}{CIFAR-100\label{fig:anytime-result:cifar100}}};
    % \node[text width=5.2cm,align=center,anchor=north] at ([yshift=-5mm]my plots c3r1.south) {\captionof{subfigure}{ImageNet\label{fig:anytime-result:imagenet}}};
\end{tikzpicture}\vspace{-5mm}
\caption{
Top-1 classification errors of anytime prediction models (lines) and fixed-budget models (isolated points) as a function of reqruied FLOPs
on (a) CIFAR-10, (b) CIFAR-100, (c) ImageNet datasets. We obtain higher or competitive accuracy of I-ResNeXt and IS-ResNeXt compared to all
other anytime and fixed-budget models, under any FLOPs or any datasets.
% Anytime prediction results. The x-axis represents required FLOPs to predict labels of one single image,
% and the y-axis does classification error (\%). The black, blue, red lines are the errors of sub-networks of I-ResNeXt, IS-ResNeXt,
% MSDNet, respectively. The isolated points are erros of single-output models, DenseNet-BC, ResNeXt, and ResNet. Note that they have
% fixed number of operations.
% I-ResNeXt and IS-ResNeXt outperform MSDNet under any FLOPs on both CIFAR datasets.
% Also they have a sub-network which achieves better performance than DenseNet-BC and ResNet models with similar FLOPs.
% Although ResNeXt has best accuracy under $1.3\times10^8$ FLOPs, I-ResNeXt and IS-ResNeXt have similar accuracy under $0.5\times10^8$ FLOPs.
}
\label{fig:anytime-result}
\end{figure}
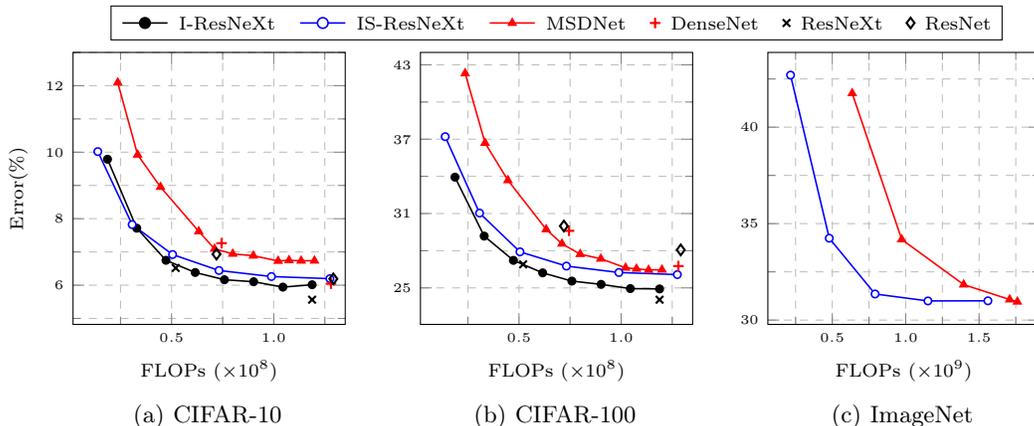

\textbf{Comparisons with fixed-budget models on CIFAR.}
We compare our anytime prediction models, 
I-ResNeXt and IS-ResNeXt, with other state-of-the-art {fixed-budget architectures}: 
ResNet \cite{resnet_he2016}, % of depth $32$ and $56$,
ResNeXt \cite{resnet_he2016} and % of depth $29$ and $65$,
DenseNet-BC \cite{densenet_huang2017} of various depths % of depth $40$ and $58$.
on the CIFAR datasets.
This is to show that jointly trained sub-networks of our models
can achieve better or competitive performance compared to solely trained, non-anytime prediction models,
i.e., to test the optimality {which} is one of desired properties for anytime prediction.
As shown in Figure \ref{fig:anytime-result:cifar10} and \ref{fig:anytime-result:cifar100},
I-ResNeXt and IS-ResNeXt have outperforming sub-networks compared to ResNet and DenseNet-BC on both CIFAR-10 and CIFAR-100 datasets.
Although ResNeXt of depth $65$ has the best performance at $1.2{\times}10^8$ FLOPs,
I-ResNeXt has a competitive sub-network at $0.5{\times}10^8$ FLOPs.

\textbf{Comparisons with MSDNet on CIFAR.}
Next, we indeed compare ours with the state-of-art anytime architecture,
MSDNet \cite{msdnet_huang2017} on CIFAR datasets.\footnote{
We use a public MSDNet model released by the authors,
available at \url{https://github.com/gaohuang/MSDNet}.
For fair comparisons, % with MSDNet,
we replace its classifiers by 1-layer classifiers as like ours.
}
Here, one might be interested in
NestedNet models \cite{nestednet_kim2017} to compare
as they also use thin sub-networks.
However, the reported performance is very poor compared to ours,
%for anytime prediction reported in \cite{nestednet_kim2017} have low accuracy,
e.g., a sub-network of the model has $24.5\%$ error on CIFAR-100, while its size is $10$ times larger
than that of I-ResNeXt of the same error.
%We also apply the NestedNet's idea to ResNet or ResNeXt models of similar FLOPs compared to ours,
%but we also obtain very poor performance of sub-networks.}
% For NestedNet, we use reported errors\footnote{We also try their approaches to ResNeXt and we obtain
% poor performance under small budgets, for example, $60\%$ errors on CIFAR-10.} in the paper \cite{nestednet_kim2017}.
As illustrated in Figure \ref{fig:anytime-result:cifar10} and \ref{fig:anytime-result:cifar100},
%both I-ResNeXt and IS-ResNeXt outperform MSDNet under any FLOPs on both CIFAR datasets.
%In particular, I-ResNeXt achieve the best performance. % on CIFAR datasets.
the full-network of I-ResNeXt has $10.7\%$, $5.8\%$ relatively smaller errors on CIFAR-10
and CIFAR-100, compared to the
full-network of MSDNet with respect to {the} same FLOPs, respectively.
The sub-networks of I-ResNeXt also have up to $56.0\%$ and $48.7\%$ smaller FLOPs compared to those of MSDNet of the same accuracy on CIFAR-10 and CIFAR-100, respectively.
Although IS-ResNeXt models are slightly worse than I-ResNeXt, they also outperforms MSDNet.
For example, the thinnest sub-network of IS-ResNeXt achieves $17.2\%$ and $12.1\%$ relatively smaller errors,
compared to 
the shallowest one of MSDNet
on CIFAR-10 and CIFAR-100, respectively, where the former has even smaller FLOPs than {the} latter.

\textbf{Comparison with MSDNet on ImageNet.}
%\textbf{ImageNet experiments.}
Finally, we train 
%As described in Section \ref{sec:method:isresnext}, 
IS-ResNeXt and MSDNet using the large-scale ImageNet dataset.
Here, we do not try I-ResNeXt for ImageNet as it takes significantly longer.\footnote{
I-ResNeXt of $1.6{\times}10^9$ FLOPs requires more than 2 weeks to train ImageNet on a Titan Xp GPU.}
%s such as ImageNet.
%To verify that our approaches can be applied under such datasets, we evaluate sub-networks of the IS-ResNeXt and compare them with MSDNet.
As shown in Figure \ref{fig:anytime-result:imagenet},
the full-network of IS-ResNeXt performs similarly to that of MSDNet, while
the sub-networks
of IS-ResNeXt have much higher accuracy than those of MSDNet.
In particular, the gap between them increases as FLOPs decreases:
sub-networks of IS-ResNeXt have up to $43.3\%$ smaller FLOPs compared to those of MSDNet of the same accuracy.
We emphasize that achieving higher performance under small FLOPs is more important for resource-limited, e.g., mobile, applications.

\subsection{Results for hierarchical anytime prediction}\label{sec:res:hanytime}

% \begin{wrapfigure}[16]{R}{0.35\textwidth}
% \centering
% %\input{NIPS2018/plots/h_anytime_cub_v2.tex}
% \vspace{-0.45cm}
% \includegraphics[width=\linewidth]{NIPS2018/plots/h_anytime_cub_v2}
% \caption{Classification errors of I-ResNeXt under hierarchical anytime prediction on CUB-200.}
% \label{fig:hanytime-result}
% \end{wrapfigure}

\begin{wrapfigure}[18]{r}{0.5\textwidth}
\centering
\begin{tikzpicture}
    \begin{groupplot}[
        group style={
            group name=anytime,
            group size=2 by 1,
            xlabels at=edge bottom,
            ylabels at=edge left,
            %vertical sep=1.7cm,
            horizontal sep=0.5cm,
        },
        height=4cm,width=4cm,
        ymajorgrids=true,yminorgrids=true,
        xmajorgrids=true,xminorgrids=true,
        minor tick num=1,
        grid style=dashed,
        yticklabel style={/pgf/number format/fixed,
                          /pgf/number format/fixed zerofill,
                          /pgf/number format/precision=0},
        xticklabel style={/pgf/number format/fixed,
                          /pgf/number format/fixed zerofill,
                          /pgf/number format/precision=1},
        tick label style = {font = \tiny},
        label style = {font = \scriptsize},
        legend style = {font = \scriptsize},
        ylabel=Error($\%$),
    ]
    \nextgroupplot[mark size=1.4pt,xlabel={FLOPs ($\times10^8$)}] % CIFAR-100
        %\addlegendentry{MSDNet}
        \addplot[mark=*,solid,line width=0.6pt,color=black]
        coordinates {
        (0.13879652, 37.204)
        (0.30708324, 31.038)
        (0.50486116, 27.908)
        (0.73213028, 26.760)
        (0.98889060, 26.242)
        (1.27514212, 26.072)
        };
        \addplot[mark=o,dashed,line width=0.6pt,color=red,mark options={solid}]
        coordinates {
        (0.13879652, 38.483)
        (0.30708324, 31.290)
        (0.50486116, 29.217)
        (0.73213028, 27.827)
        (0.98889060, 26.930)
        (1.27514212, 26.513)
        };
        \addplot[mark=square,dashed,line width=0.6pt,color=red,mark options={solid}]
        coordinates {
        (0.13879652, 26.840)
        (0.30708324, 21.103)
        (0.50486116, 19.270)
        (0.73213028, 18.017)
        (0.98889060, 17.270)
        (1.27514212, 16.960)
        };
        \addplot[mark=*,solid,line width=0.6pt,color=red]
        coordinates {
        (0.13879652, 26.840)
        (0.30708324, 21.103)
        (0.50486116, 19.270)
        (0.73213028, 27.827)
        (0.98889060, 26.930)
        (1.27514212, 26.513)
        };
        \coordinate (top) at (rel axis cs:0,1);
     
    \nextgroupplot[mark size=1.4pt,xlabel={FLOPs ($\times10^9$)},legend pos=outer north east] % CUB-200
    %\addlegendentry{fine-level only}
        \addplot[mark=o,dashed,line width=0.6pt,color=red,mark options={solid}]
        coordinates {
        (0.184415464, 68.309)
        (0.407365096, 59.413)
        (0.668849896, 54.139)
        (0.968869864, 50.638)
        (1.307425000, 49.644)
        (1.684515304, 48.899)
        };\label{plots:h:d1}%\addlegendentry{$d=1$}
        
        \addplot[mark=square,dashed,line width=0.6pt,color=red,mark options={solid}]
        coordinates {
        (0.184415464, 64.411)
        (0.407365096, 55.430)
        (0.668849896, 50.369)
        (0.968869864, 47.166)
        (1.307425000, 46.051)
        (1.684515304, 45.368)
        };\label{plots:h:d2}%\addlegendentry{$d=2$}
        
        \addplot[mark=triangle,dashed,line width=0.6pt,color=red,mark options={solid}]
        coordinates {
        (0.184415464, 61.229)
        (0.407365096, 52.271)
        (0.668849896, 47.314)
        (0.968869864, 44.104)
        (1.307425000, 42.920)
        (1.684515304, 42.282)
        };\label{plots:h:d3}%\addlegendentry{$d=3$}
        
        \addplot[mark=diamond,dashed,line width=0.6pt,color=red,mark options={solid}]
        coordinates {
        (0.184415464, 51.888)
        (0.407365096, 44.011)
        (0.668849896, 39.734)
        (0.968869864, 37.014)
        (1.307425000, 35.623)
        (1.684515304, 34.850)
        };\label{plots:h:d4}%\addlegendentry{$d=4$}
        
        \addplot[mark=*,solid,line width=0.6pt,color=black]
        coordinates {
        (0.184415464, 68.760)
        (0.407365096, 61.302)
        (0.668849896, 53.614)
        (0.968869864, 50.692)
        (1.307425000, 48.942)
        (1.684515304, 48.994)
        };\label{plots:h:fine}%\addlegendentry{fine-only}
        
        \addplot[mark=*,solid,line width=0.6pt,color=red]
        coordinates {
        (0.184415464, 51.674)
        (0.407365096, 52.507)
        (0.668849896, 50.600)
        (0.968869864, 50.768)
        (1.307425000, 49.823)
        (1.684515304, 49.072)
        };\label{plots:h:joint}%\addlegendentry{joint}
        
    % \nextgroupplot[mark size=1.4pt,xlabel={FLOPs ($\times10^9$)}] % Dogs
    %     \addplot[mark=o,dashed,line width=0.6pt,color=red,mark options={solid}]
    %     coordinates {
    %     (0.184415464, 55.513)
    %     (0.407365096, 43.310)
    %     (0.668849896, 42.343)
    %     (0.968869864, 40.991)
    %     (1.307425000, 41.818)
    %     (1.684515304, 41.737)
    %     };
        
    %     \addplot[mark=square,dashed,line width=0.6pt,color=red,mark options={solid}]
    %     coordinates {
    %     (0.184415464, 54.056)
    %     (0.407365096, 42.168)
    %     (0.668849896, 40.455)
    %     (0.968869864, 39.371)
    %     (1.307425000, 40.058)
    %     (1.684515304, 40.093)
    %     };
        
    %     \addplot[mark=triangle,dashed,line width=0.6pt,color=red,mark options={solid}]
    %     coordinates {
    %     (0.184415464, 41.119)
    %     (0.407365096, 30.431)
    %     (0.668849896, 28.811)
    %     (0.968869864, 28.159)
    %     (1.307425000, 27.774)
    %     (1.684515304, 27.984)
    %     };
        
    %     \addplot[mark=diamond,dashed,line width=0.6pt,color=red,mark options={solid}]
    %     coordinates {
    %     (0.184415464, 27.517)
    %     (0.407365096, 20.641)
    %     (0.668849896, 19.138)
    %     (0.968869864, 18.706)
    %     (1.307425000, 18.858)
    %     (1.684515304, 18.811)
    %     };
        
    %     \addplot[mark=*,solid,line width=0.6pt,color=black]
    %     coordinates {
    %     (0.184415464, 66.010)
    %     (0.407365096, 52.030)
    %     (0.668849896, 42.480)
    %     (0.968869864, 39.590)
    %     (1.307425000, 39.340)
    %     (1.684515304, 39.280)
    %     };
        \coordinate (bot) at (rel axis cs:1,0);
    \end{groupplot}
    legend
    \path (top|-current bounding box.north)--
          coordinate(legendpos)
          (bot|-current bounding box.north);
    \matrix[
        matrix of nodes,
        font=\scriptsize,
        anchor=south,
        draw,
        inner sep=0.2em,
        draw
    ]at([yshift=1ex]legendpos)
    {
        \ref{plots:h:d1}    & $d=1$     &[5pt]
        \ref{plots:h:d3}    & $d=3$     &[5pt]
        \ref{plots:h:fine}  & fine-only &[5pt]
        \\ 
        \ref{plots:h:d2}    & $d=2$     &[5pt]
        \ref{plots:h:d4}    & $d=4$     &[5pt]
        \ref{plots:h:joint} & coarse$\rightarrow$fine
        \\
    };
    \node [text width=4cm,anchor=north,align=center] at ([yshift=-7mm]anytime c1r1.south) {\subcaption{CIFAR-100\label{fig:hanytime-result:cifar100}}};
    \node [text width=4cm,anchor=north,align=center] at ([yshift=-7mm]anytime c2r1.south) {\subcaption{CUB-200\label{fig:hanytime-result:cub200}}};
    
    % \node [text width=5.2cm,anchor=north,align=center] at ([yshift=-7mm]anytime c3r1.south) {\subcaption{Dogs\label{fig:hanytime-result:dogs}}};
    % \node[text width=5.2cm,align=center,anchor=north] at ([yshift=-5mm]my plots c1r1.south) {\captionof{subfigure}{CIFAR-10\label{fig:anytime-result:cifar10}}};
    % \node[text width=5.2cm,align=center,anchor=north] at ([yshift=-5mm]my plots c2r1.south) {\captionof{subfigure}{CIFAR-100\label{fig:anytime-result:cifar100}}};
    % \node[text width=5.2cm,align=center,anchor=north] at ([yshift=-5mm]my plots c3r1.south) {\captionof{subfigure}{ImageNet\label{fig:anytime-result:imagenet}}};
\end{tikzpicture}\vspace{-5mm}
\caption{
Top-1 classification errors of IS-ResNeXt for hierarchical anytime prediction on (a) CIFAR-100 and (b) CUB-200.
%,(c) Dogs datasets.
\iffalse
Each dashed line represents anytime prediction errors of a certain abstraction-level
labels. The red solid lines show how to maintain a certain accuracy level across all FLOPs.
The black solid lines are results of training fine labels only.
\fi
% Anytime prediction results. The x-axis represents required FLOPs to predict labels of one single image,
% and the y-axis does classification error (\%). The black, blue, red lines are the errors of sub-networks of I-ResNeXt, IS-ResNeXt,
% MSDNet, respectively. The isolated points are erros of single-output models, DenseNet-BC, ResNeXt, and ResNet. Note that they have
% fixed number of operations.
% I-ResNeXt and IS-ResNeXt outperform MSDNet under any FLOPs on both CIFAR datasets.
% Also they have a sub-network which achieves better performance than DenseNet-BC and ResNet models with similar FLOPs.
% Although ResNeXt has best accuracy under $1.3\times10^8$ FLOPs, I-ResNeXt and IS-ResNeXt have similar accuracy under $0.5\times10^8$ FLOPs.
}
\label{fig:hanytime-result}
\vspace{-0.3cm}
\end{wrapfigure}
We use the CIFAR-100 {and} CUB datasets to evaluate the performance
of IS-ResNeXt for hierarchical anytime prediction, where the coarse labels of them are obtained from their taxonomies
built by WordNet \cite{wordnet_miller1995} (see \cite{eval_fine_akata2015}).
The details about the taxonomies are described in the supplementary material.
%Appendix \ref{sec:appendix:taxonomy}. 
All sub-networks of IS-ResNeXt are jointly trained under the hierarchical anytime loss \eqref{eq:hanytime-loss}.
% We only choose three coarse labels for each fine label, i.e., set $D=4$. We train $K=6$ sub-networks
% of IS-ResNeXt jointly under
% minimizing the anytime loss \eqref{eq:hanytime-loss}. The IS-ResNeXt has $L=16$ residual blocks of
% width $W=192$.
The results are illustrated in Figure \ref{fig:hanytime-result}.
As expected, the finer labels are more difficult to classify and larger models can
classify labels more accurately. By choosing appropriate coarse and fine labels targeted by
small and large sub-networks, respectively,
IS-ResNeXt maintains its performance at a certain level across all sub-networks.
For example, all sub-networks can have at most $52.5\%$ error under
any budgets on the CUB dataset, while 
the error increases up to $68.8\%$ without using coarse labels.
At an angle, this result might not be too surprising as we improve the performance by losing the predictive information.
However, the modified anytime prediction task is useful to match the original, usable accuracy in practical applications and
it is remarkable that the performance of IS-ResNeXt is not degraded even under significantly more auxiliary
classifiers attached.

% Each red dashed lines represent classification errors of each coarse-level.
% When the $k=1,2,3$-th sub-networks predict the coarse labels of $d=1,2,3$, respectively,
% and other sub-networks predict the fine label of $d=4$, the IS-ResNeXt under \eqref{eq:hanytime-loss} 
% achieves $53.0\%$ maximum errors anytime (red solid line).
% However, without considering coarse labels, i.e., under the original anytime loss \eqref{eq:anytime-loss},
% IS-ResNeXt obtains up to $68.8\%$ errors when small budgets are only allowed.

\section{Conclusion}
We aim for claiming that utilizing thin sub-networks of the same depth is more effective than shallow ones for anytime prediction. 
By focusing recent state-of-the-art multi-branch networks,
we propose new models that outperforms prior ones.
%To build thin sub-networks, we first propose I-ResNeXt which has thin sub-networks by removing branches from a multi-branch architecture.
%Moreover, to handle the inefficiency of joint-training the thin sub-networks, we propose a novel architecture, IS-ResNeXt, which can be
%jointly trained with their sub-networks efficiently.
%These two architectures outperform existing anytime or fixed-budget architectures on CIFAR and ImageNet datasets.
We also propose a new anytime prediction problem, for maintaining performance
across anytime budgets via allowing rough predictions.
%We apply IS-ResNeXt to this problem and achieve the desired goal, maintaining performance under anytime budgets.
%Since our generic ideas are applicable to any multi-branch architectures,
We hope that our results would be beneficial to many related applications or problems in the future.

% This paper provides experimental supports rediscovering that many recent deep architectures are potentially
% able to perform multi-operations easily, by building many inclusive thin sub-networks of the same depth.
% Under residual net- works on CIFAR datasets, we found that the performance under the joint training is
% not degraded or often improved although it can perform multi-operations or have fewer parameters.
% We hope that our results would be beneficial to many related applications or problems in the future.

\subsubsection*{Acknowledgments}
We would like to acknowledge Jongheon Jeong and Kimin Lee for helpful discussions,
and Kibok Lee for providing the hierarchical taxonomy of the Caltech-UCSD Birds dataset.
The research was funded by Naver Labs.
%This work was funded by Naver Labs.

\small
\printbibliography

\appendix

\section{Training setups}\label{sec:appendix:setup}
% the \\ insures the section title is centered below the phrase: AppendixA

\textbf{Datasets.}
We use CIFAR-10/100 \cite{cifar_krizhevsky2009a},
ImageNet \cite{imagenet_deng2009} and
Caltech-UCSD Birds (CUB) \cite{cub_wah2011} datasets.
%,and Stanford Dogs \cite{dog_khosla2011} datasets.
%hierarchical anytime prediction.
The CIFAR datasets have $50{,}000$ training
images and $10{,}000$ test images. Each image has $32^2$ size and each pixel has RGB color. CIFAR-10/100 
have 10 and 100 classes, respectively.
% Each class in CIFAR-10 has $5{,}000$ training images and $1{,}000$ test images.
% Similarly, in CIFAR-100, each class has $500$ and $100$ images in training and test sets, respectively.
Following \cite{resnet_he2016}, we use data-augmentation techniques to the training images: random horizontal flip,
random-crop with 4 pixel zero-padding, normalizing pixel value by channel means and standard deviations.
ImageNet has
$1.2$ million training images and $50{,}000$ validation images of $1,000$ classes.
Similarly, CUB has $5{,}994$ training images and $5{,}794$ test images of 200 fine-grained bird species.
For hierarchical anytime prediction, we obtain a hierarchical taxonomy of the CUB dataset from WordNet \cite{wordnet_miller1995} by following \cite{eval_fine_akata2015}.
We obtain $99$ non-leaf nodes (i.e., coarse labels) from the taxonomy with maximum depth $8$.
By taking ancestors (i.e., coarse labels) of the leaf nodes (i.e., original fine-grained labels) up to distance $3$,
we build $D=4$ different levels of labels: $200$, $183$, $149$ and $80$ labels from fine-grained to coarse-grained.
%Dogs also has $12{,}000$ training images, $8{,}580$ test images, and 120 fine-grained dog labels.
Note that CIFAR-100 has its own $20$ coarse-grained labels.
We use the same augmentation techniques as \cite{resnet_he2016} for training ImageNet and CUB images.
Note that the image size after the augmentations is $224^2$.

\textbf{Optimization.}
All models are trained by stochastic gradient descent (SGD) with 
Nesterov momentum of momentum $0.9$ without dampening and MSRA initialization \cite{msra_he2015}.
We use a weight decay of $10^{-4}$ and an initial learning rate of $0.1$ for all experiments.
% For CIFAR and CUB datasets, we use a batch size of $64$,
% a batch size of $64$ and an initial learning rate of $0.1$.
% We use a weight decay of $10^{-4}$, a batch size of $64$ and an initial learning rate of $0.1$.
The models for CIFAR and CUB are trained for $300$ and $150$ epochs, respectively, with a batch size of $64$.
The learning rate is divided by 10 after $50\%$ and $75\%$ epochs.
For ImageNet, we use same hyperparameters as CIFAR except the learning schedule, where the total number of epochs is $90$ and
learing rate is diveded at $30$ and $60$ epochs, and the batch size of $96$.
We randomly select $5{,}000$, $50{,}000$, $1{,}000$ images of the training set for validation in CIFAR, ImageNet,
and CUB datasets, respectively.
Note that we use the original validation set as test set in ImageNet. All models are averaged on 5 trials.

\section{Details on model architectures}
For I-ResNeXt and IS-ResNeXt, we denote the width of features (e.g., $\mathbf{x},\mathbf{y}$) as $W$,
the width of bottleneck features (e.g., $\mathbf{a},\mathbf{b}$) as $B$, cardinality (or the number of branches) as $C$,
the number of sub-networks as $K$ and the number of blocks as $L$.
Similarly to ResNet-based architectures, we use the same number of blocks for each scale.
For down-scaling, average pooling layers of stride $2$ are used with adding zero-filled features for increasing width
as like the type-A shortcut connection in \cite{resnet_he2016}.
% Features are
% down-sampled by an average pooling layer with stride 2
% and zero-padding for increasing width as like the type-A
% shortcut connection in \cite{resnet_he2016}.
Note that the bottleneck width is also doubled like other ResNet models.
% When the scale is halved, the widths $W,B$ are doubled. For down-scaling and expanding width, we use average pooling
% with a stride of 2 and padding zero-filled channels.
We use 3 scales for CIFAR, and 4 scales for other datasets.
It means that models for CIFAR have $32^2,16^2,8^2$-sized features of $W,2W,4W$ channels, respectively;
models for ImageNet and CUB similarly have $56^2,28^2,14^2,7^2$-sized features of $W,2W,4W,8W$ channels, respectively.
For all models, we use $B=4$ as width of bottleneck in the first scale.
Before the first block, we use a $3{\times}3$ convolutional layer
for CIFAR images; and a $7{\times}7$ convolutional layer of stride $2$ and a max pooling layer of kernel size $2$
for ImageNet and CUB. We also set the cardinality
by $C=0.5W/B$ and $C=0.75W/B$ for I-ResNeXt and IS-ResNeXt, respectively.
Each classifier consists of a BN layer, a ReLU activation, a global average pooling layer, and a fully-connected layer.
Namely, the total depth of the architectures is $3L+2$.
We remind again that that all convolutional weights are shared among all sub-networks.
% Note that we replace 3-layer classifiers by
% 1-layer linear classifiers in MSDNet \cite{msdnet_huang2017} for fair comparisons.

We use different hyperparameters for $L, W, K$ depending on datasets and tasks.
For Section 4.1 in the main paper, we use small I-ResNeXt models of $L=9$ and $W=64$ and build $K=8$ thin sub-networks.
In other sections, we use different hyperparameters depending on only datasets:
I-ResNeXt models of $L=21$, $W=64$, $K=8$ and IS-ResNeXt of $L=15$, $W=96$, $K=6$ for CIFAR;
IS-ResNeXt of $L=20$, $W=160$, $K=5$ for ImageNet;
IS-ResNeXt of $L=16$, $W=192$, $K=6$ for CUB.

\end{document}